\documentclass[lettersize,journal]{IEEEtran}
\usepackage{amsmath,amsfonts}
\usepackage{algorithmic}
\usepackage{algorithm}
\usepackage{array}
\usepackage[caption=false,font=normalsize,labelfont=sf,textfont=sf]{subfig}
\usepackage{textcomp}
\usepackage{stfloats}
\usepackage{url}
\usepackage{verbatim}
\usepackage{graphicx}
\usepackage{cite}
\usepackage{multirow}
\usepackage{booktabs}
\usepackage{tabularx}
\usepackage[table]{xcolor}
\usepackage{xpatch}
\usepackage[section]{placeins}
\usepackage[colorlinks=black,linkcolor=black]{hyperref}

\makeatletter
\def\changeBibColor#1{%
	\in@{#1}{li2022nested, xiong2024deep, baik2024dbn}
	\ifin@\color{black}\else\normalcolor\fi
}

\xpatchcmd\@bibitem
{\item}
{\changeBibColor{#1}\item}

\xpatchcmd\@lbibitem
{\item}
{\changeBibColor{#2}\item}

\makeatother

\begin{document}

\title{GroupFace: Imbalanced Age Estimation Based on  Multi-hop Attention Graph Convolutional Network and Group-aware Margin Optimization}

\author{IEEE Publication Technology,~\IEEEmembership{Staff,~IEEE,}
\author{Yiping~Zhang, Yuntao~Shou, Wei~Ai, Tao~Meng, and~Keqin~Li,~\IEEEmembership{Fellow,~IEEE}
	\thanks{Corresponding Author: Tao Meng~(mengtao@hnu.edu.cn)}
	
	\IEEEcompsocitemizethanks{\IEEEcompsocthanksitem Y. Zhang, ~Y. Shou, ~W. Ai, and ~T. Meng are with College of Computer and Mathematics, Central South University of Forestry and Technology, Changsha, Hunan 410004, China. (yipingzhang@csuft.edu.cn, shouyun-tao@stu.xjtu.edu.cn, aiwei@hnu.edu.cn,~mengtao@hnu.edu.cn)
	\IEEEcompsocthanksitem K. L is with the Department of Computer Science, State University of New York, New Paltz, New York 12561, USA. (lik@newpaltz.edu)}}

\thanks{This paper was produced by the IEEE Publication Technology Group. They are in Piscataway, NJ.}
\thanks{Manuscript received April 19, 2021; revised August 16, 2021.}}

\markboth{Journal of \LaTeX\ Class Files,~Vol.~14, No.~8, August~2021}%
{Shell \MakeLowercase{\textit{et al.}}: A Sample Article Using IEEEtran.cls for IEEE Journals}


\maketitle

\begin{abstract}
With the recent advances in computer vision, age estimation has significantly improved in overall accuracy. However, owing to the most common methods do not take into account the class imbalance problem in age estimation datasets, they suffer from a large bias in recognizing long-tailed groups. To achieve high-quality imbalanced learning in long-tailed groups, the dominant solution lies in that the feature extractor learns the discriminative features of different groups and the classifier is able to provide appropriate and unbiased margins for different groups by the discriminative features. Therefore, in this novel, we propose an innovative collaborative learning framework (GroupFace) that integrates a multi-hop attention graph convolutional network and a dynamic group-aware margin strategy based on reinforcement learning. Specifically, to extract the discriminative features of different groups, we design an enhanced multi-hop attention graph convolutional network. This network is capable of capturing the interactions of neighboring nodes at different distances, fusing local and global information to model facial deep aging, and exploring diverse representations of different groups. In addition, to further address the class imbalance problem, we design a dynamic group-aware margin strategy based on reinforcement learning to provide appropriate and unbiased margins for different groups. The strategy divides the sample into four age groups and considers identifying the optimum margins for various age groups by employing a Markov decision process. Under the guidance of the agent, the feature representation bias and the classification margin deviation between different groups can be reduced simultaneously, balancing inter-class separability and intra-class proximity. After joint optimization, our architecture achieves excellent performance on several age estimation benchmark datasets. It not only achieves large improvements in overall estimation accuracy but also gains balanced performance in long-tailed group estimation.
\end{abstract}


\section{Introduction}
\IEEEPARstart{A}{ge} is one of the most important biometric traits in faces, and the general precision in age estimation has seen notable advancements in recent years. It encompasses a broad spectrum of application scenarios, including social media, visual surveillance, image retrieval, marketing, and public safety \cite{rothe2015dex, zhang2020ssr, wang2023exploiting, shou2022conversational, shou2025masked, shou2022object, shou2023comprehensive, shou2024adversarial, meng2024deep, shou2023adversarial}.

\begin{figure}[!t]
	\centering
	\includegraphics[width=0.48\textwidth]{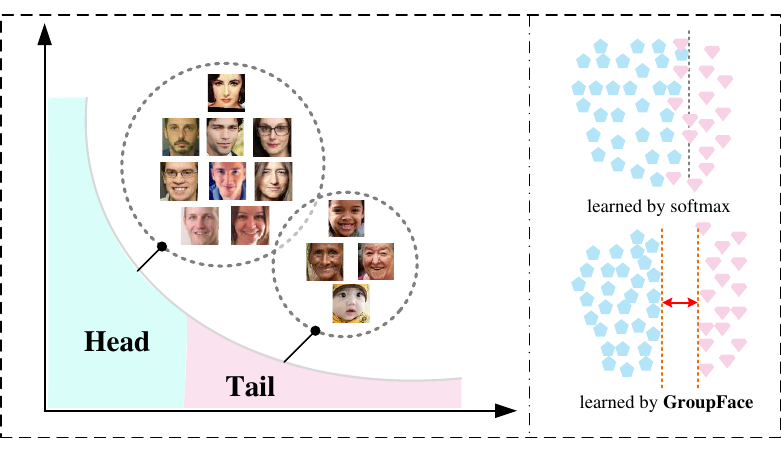}
	\caption{The illustration of age estimation with class imbalanced learning. Most face datasets have an imbalanced distribution of race and age groups, such that the recognition bias is high for the long-tailed groups. Our GroupFace can achieve a balanced generalization capability for different age groups by discriminative feature extraction and group-aware margin optimization.}
	\label{fig_1}
\end{figure}

Driven by the rapid development of deep learning, age estimation based on Convolutional Neural Networks (CNN) achieves promising performance. Rothe \emph{et al.} \cite{rothe2015dex} utilized a pre-trained VGG-16 network for face representation learning, then used the classification probability multiply the corresponding labels to get a regression result, which is much better than the single classification or regression methods. Zhang \emph{et al.} \cite{chen2017using} proposed Ranking-CNN, which converted the age estimation task into a ranking challenge, and obtained age prediction results by summing the binary classification results. With the advent of Vision Transformer (ViT) \cite{dosovitskiy2020image}, its powerful ability to globally model dependencies was put into use for face representation learning. 
Kuprashevich \emph{et al.} \cite{kuprashevich2023mivolo} proposed a unified dual model Multi Input VOLO (MiVOLO), which integrated the gender and age estimation tasks in the field based on the Vision Transformer. Qin \emph{et al.} \cite{qin2023swinface} proposed Transformer-based SwinFace, which achieved multi-task face feature extraction such as age estimation through a shared backbone and a sub-net for each related task. Moreover, benefits from the flexibility of Graph Convolutional Neural Networks (GCN) \cite{ai2023gcn, ai2024gcn, meng2024multi, shou2024contrastive, shou2024spegcl, shou2024efficient, ying2021prediction, shou2023graph, shou2023czl, meng2024masked, shou2024revisiting, ai2024edge, zhang2024multi} in processing complex and irregular objects, GCN achieves performance similar to or even surpassing that of Transformer in feature extraction. Shou \emph{et al.} \cite{shou2025masked} proposed Masked Contrastive Graph Representation Learning (MCGRL) to capture the rich structural face information more flexibly and with low redundancy, which outperformed most of the CNN and ViT-based methods on age estimation task. These frontier age estimation researches mainly focus on designing feature extraction networks for more robust face representation learning and optimizing age estimation strategies for more accurate prediction performance, which have achieved great improvement in the overall age estimation performance. However, a prominent issue is that these approaches overlook the issue of \textcolor{black}{imbalanced} data distribution, fail to maintain balanced performance on long-tailed group recognition and show significant degradation when encountering specific scenarios.

Due to different difficulties in collecting samples from different age groups, the age estimation datasets imbalance problem is so severe that the prediction accuracy for some groups or specific application scenarios is poor. For example, in the long-span dataset, MIVIA \cite{greco2022effective}, the number of images for certain age groups (0.6$\%$ for ``children", 4.4$\%$ for ``teenager", and 6.3$\%$ for ``senior") is much lower than for the head age group (88.7$\%$ for ``adult"), and this imbalanced distribution leads to biased identification of long-tailed groups. In many cases, although the tail categories are numerical minorities, ignoring them will be costly. Application scenarios such as grading systems in social media, anti-addiction of minors in games, searching for lost children in public safety, and fraud prevention for the elderly. Seldom studies have focused on long-tailed recognition in age estimation. Bao \emph{et al.} \cite{bao2021lae} designed a two-stage framework that decouples the learning process into representation learning and classification, which utilized a balanced sampling dataset to train a balanced classifier. Deng \emph{et al.} \cite{deng2021pml} utilized globally-tuned ResNet-34 for feature extraction and proposed the variational margin to minimize the effect of misleading tail sample predictions for head class. Wang \emph{et al.} \cite{wang2023exploiting} proposed Meta-Set Learning (MSL) based on RestNet-34 and created an unfair filtering network to identify and filter the noisy samples, and then obtained a balanced meta-set for meta-weighting to alleviate the unfairness between age estimation. Though these techniques mitigate the class imbalance issue to some extent, they still have some shortcomings: \textbf{i) Weak in discriminative feature learning.} The CNN-based local feature extraction \cite{deng2021pml, wang2023exploiting} or Transformer-based global feature extraction \cite{kuprashevich2023mivolo, qin2023swinface} tends to learn the consistency of most classes, which is difficult to effectively learn the personalized information of different samples. In particular, the lack of discriminative sample mining leads to \textcolor{black}{the over-fitting} of minority classes. \textbf{ii) Poor in identifying appropriate and unbiased margins for different groups.} Some methods \cite{shou2025masked, bao2021lae} decoupled the two-stage learning process, which is no explicit control over the distribution of learned features and hard to provide optimal margins for different groups from staged extracted features. Other methods \cite{cai2022meta, wang2023exploiting} introduced meta-learning for learning the distribution of imbalanced samples, but meta-learning is extremely sensitive to noise, making it difficult to flexibly and stably provide appropriate margins for different groups.

To overcome the above shortcomings for improving the generalization performance of age estimation, it is necessary to focus on both discriminative feature extraction of different groups and \textcolor{black}{an imbalanced} margin learning strategy for long-tailed classes. In this novel, an innovative collaborative learning framework (GroupFace) is proposed that integrates a multi-hop attention graph convolutional network and dynamic group-aware margin strategy based on reinforcement learning. To extract the discriminative features among different age groups, we design an enhanced multi-hop attention graph convolutional network. It has the capability to capture the interactions of neighboring nodes at different distances, effectively extending the receptive field of the graph model. It also models facial deep aging by fusing local and global information, thus capturing diverse representations of different groups. Specifically, to maintain message diversity for deeper mining of discriminative features, we design an adaptive decay strategy during graph diffusion to adaptively assign learnable weights based on different hop distances, randomly drop messages during message propagation to prevent over-fitting and incorporate residual blocks in graph convolution to prevent over-smoothing. Moreover, to address the computational inefficiency of higher-order graph models, we utilize a power iteration method to approximate the accelerated inverse matrix.

At the same time, to further address the class imbalance problem, we design a dynamic group-aware margin strategy based on reinforcement learning. We roughly categorize ages into four groups (``children", ``teenager", ``adult", and ``senior"), and design a more flexible and stable group-aware margin loss function. While considering employing a Markov decision process to identify the optimum margins for various age groups, we utilize deep q-learning to acquire a strategy for selecting appropriate group margins. As shown in Fig. \ref{fig_1}, under the guidance of the agent, the representation bias in the feature space and the margin deviation in the classification space between different groups can be reduced simultaneously, while the inter-class separability and intra-class proximity be improved, then achieving a balanced generalization ability. The resulting adaptive and unbiased margins for different age groups are more conducive to subsequent accurate age classification and regression from coarse to fine.

Finally, through joint optimization, the group-aware margin policy will guide and facilitate our whole age estimation architecture GroupFace. The contributions to this novel are concluded as follows:

\begin{itemize}
	\item
	We propose an innovative collaborative learning framework (GroupFace) that integrates a multi-hop attention graph convolutional network and a group-aware margin strategy, which focuses on both discriminative feature extraction of different groups and an imbalanced margin learning strategy for long-tailed classes.
	
	\item
	To achieve more discriminative representation learning, we propose an enhanced multi-hop attention graph convolutional network fusing local and global information to model aging changes in faces and design adaptive decay diffusion, random message dropping, and power iteration methods to enhance the graph model.
	
	\item
    To overcome the imbalanced distribution problem, we propose a dynamic group-aware margin strategy based on reinforcement learning with a more flexible and stable margin loss function. The balanced generalization ability is achieved by facilitating the search \textcolor{black}{for} optimal margins for different age groups through reinforcement learning.
    
   	\item 
   	Extensive experiments have shown that our architecture not only provides a significant improvement in overall estimation accuracy but also balances performance in long-tailed groups.
\end{itemize}

\section{Related Work}
In this section, we first present an overview of the frontier research on Graph Neural Networks. Then we review some studies in the general and imbalanced age estimation. Finally, we provide a brief description of reinforcement learning.

\subsection{Graph Neural Network}
Graph Neural Networks and their variants extend deep networks from regular grids to non-Euclidean graph-structured data, showing great potential in areas such as action recognition \cite{ahmad2021graph}, social media \cite{yan2024feature}, traffic prediction \cite{li2023dynamic} and computer vision \cite{zhang2024multi}.

Existing models mainly adhere to a message-passing structure, aggregating information from adjacent nodes in a direct connection. Graph Convolutional Neural Network (GCN) \cite{kipf2016semi, ai2023two, meng2024revisiting, ai2024mcsff, shou2023graphunet, ai2024graph, shou2024low, shou2024graph, ai2024seg, ai2024contrastive, fu2024sdr, shou2024dynamic} aggregated the representations of their one-hop neighbor nodes in a recursive manner and in the process normalized the weights of the edges using the Laplace matrix. On the other hand, Graph Attention Network (GAT) \cite{velivckovic2017graph} introduced a multi-head self-attention mechanism that dynamically discerned the significance of various neighboring nodes, thus discarding the fixed neighbor weight settings in traditional methods. To address the challenges of processing large graph data, GraphSAGE\cite{hamilton2017inductive} proposed an efficient batch training algorithm that ensures constant computational and memory complexity regardless of the graph size, which significantly improved the scalability of large graphs. However, these methods restrict information extraction to the local neighborhood, limiting deep feature extraction and representation of the graph model.

Since the use of multiple single-hop message-passing layers might lead to a decline in model efficacy due to the effects of Laplace smoothing, multi-hop graph networks have been put forth to address this issue by capturing information from the k-hop neighborhood vicinity. Simplified Graph Convolutional Network (SGC) \cite{wu2019simplifying} utilized powers of the adjacency matrix to generate multi-hop neighbor representations. Abu-El-Haija \emph{et al.} \cite{abu2019mixhop} designed MixHop, a mixed-hop GNN that broadens the receptive field by reiterating the feature representations of neighbors at varying distances. Wang \emph{et al.} \cite{wang2020multi} proposed Multi-hop Attention Graph Neural Network (MAGNA) to deal with the over-smoothing problem by graph attention and diffusion methods, but fixed decay weights make it inflexible for different nodes at the same distance. In this work, we pay attention to the optimization of \textcolor{black}{the multi-hop model} to obtain more discriminative face features.

\subsection{General Age Estimation}
Age estimation becomes a challenging task due to many internal and external factors, and the mainstream works for general age estimation have focused on two directions: designing more robust feature extraction models \cite{zhang2019c3ae, shou2025masked} and optimizing more accurate age prediction strategies \cite{shin2022moving, chen2023daa}. Zhang \emph{et al.} \cite{zhang2019c3ae} proposed C3AE to train multi-scale images using cascade networks to make full use of contextual information. Shin \emph{et al.} \cite{shin2022moving} developed a new sequential regression technique named Moving Window Regression (MWR), integrating the notion of relative rank and establishing both local and global relative regressors to attain the rho-rank in the whole and specific rank ranges. Chen \emph{et al.} \cite{chen2023daa} presented the Delta Age AdaIN (DAA), comprising components such as a facial encoder, DAA operation, binary code mapping, and age decoder module. Shou \emph{et al.} \cite{shou2025masked} proposed the new representation framework Masked Contrastive Graph Representation Learning (MCGRL), which can learn face structure and semantics flexibly.

These techniques designed for general age estimation tasks have enhanced the overall precision. However, overlooking the long-tailed distribution of age datasets, their performance deteriorates severely when it comes to younger and older individuals.

\textcolor{black}{
\subsection{Imbalanced Age Estimation}
The imbalanced distribution of data is widespread in real life. There has been quite a bit of groundwork in the visual imbalanced learning species to address such problems, which can be categorized into two main types: data level \cite{bao2021lae, xiong2024deep} and algorithm level \cite{li2022nested, baik2024dbn}. The data level
balances the class distribution by resampling the training data, but this tends to destroy the original expression. The algorithm level improves the importance of minority classes by improving existing algorithms, including cost-sensitive learning, ensemble learning, and other methods.}

\textcolor{black}{
However, studies dealing with imbalanced age estimation are still scarce, and ignoring minorities is disastrous and costly in some scenarios. Bao \emph{et al.} \cite{bao2021lae} decoupled face representation learning and age classification at the data level by training a balanced classifier separately from a balanced dataset obtained by class-balanced sampling. Deng \emph{et al.} \cite{deng2021pml} proposed variational margins at the algorithm level to mitigate the misleading tail-sample prediction in the head class. Wang \emph{et al.} \cite{wang2020mitigating} carefully created an unfair filtering network under a meta-learning paradigm that utilizes meta-re-weighting interventions to reduce training bias caused by category imbalance. Bao \emph{et al.} \cite{bao2023general} proposed Pixel-level Auxiliary learning (PA) and Feature Rearrangement (FR) to better utilize the facial features, while Adaptive Routing (AR) was devised to select the appropriate classifiers to improve the long-tailed recognition.}

In this work, we consider both the discriminative \textcolor{black}{feature} extraction and the imbalanced learning to achieve a balanced performance for the long-tailed groups while improving the overall accuracy.

\subsection{Reinforcement Learning}
Reinforcement learning mimics the human decision-making process training agents to learn trial-and-error-based strategies by maximizing cumulative rewards in dynamic environments. In addition to applications such as robot control and gaming, reinforcement learning has recently been successfully applied to several visual recognition tasks. Lin \emph{et al.} \cite{lin2020deep} modeled a sequential decision-making process to classify the images, and devised more rewards for the minority category. Liu \emph{et al.} \cite{liu2019fair} presented fair loss, which is a margin-aware reinforcement learning-based loss function to learn an adaptive margin. Wang \emph{et al.} \cite{wang2020mitigating} proposed a reinforcement learning-based Racial Balancing Network (RL-RBN)to find the most suitable margin for non-Caucasians and can reduce the skewness of feature dispersion among races. In this work, a dynamic group-aware margin strategy based on reinforcement learning is designed with a more flexible and stable margin loss function for imbalanced age estimation.

\section{Preliminaries}
\begin{figure*}[!t]
	\centering
	{\includegraphics[width=1\textwidth]{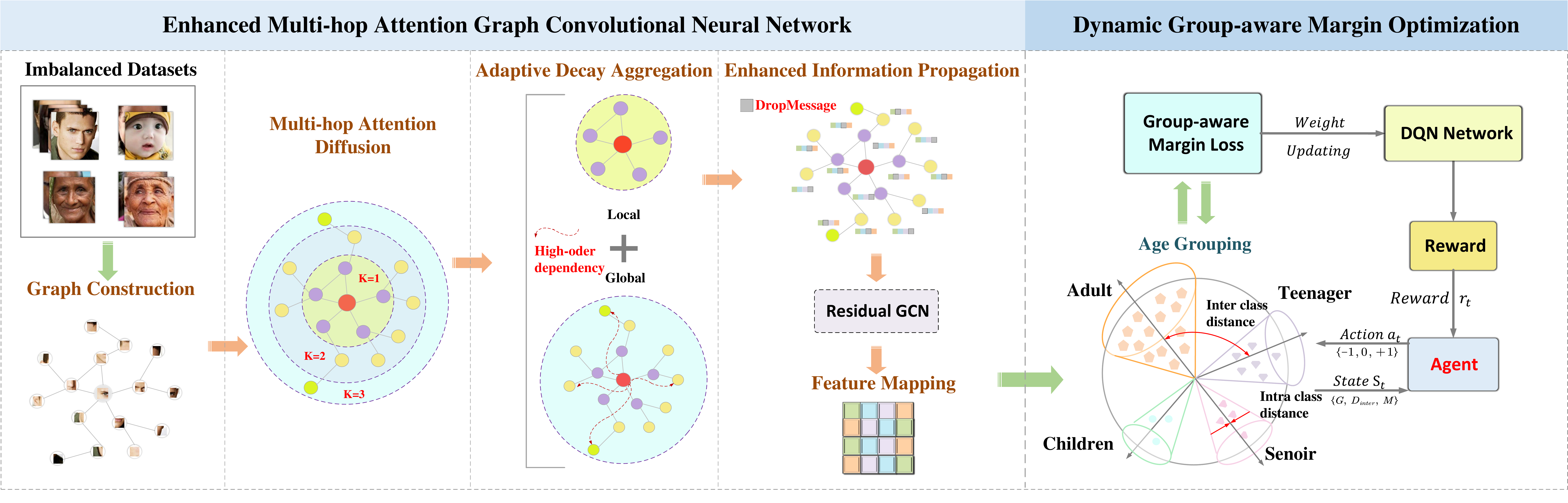}}
	\caption{\textbf{The overall framework} of our proposed imbalanced learning method \textbf{GroupFace}. Face images are segmented into patches as nodes, and then a multi-hop attention graph convolutional network will fuse global and local information to model deep facial aging for capturing the discriminative features of different groups. Through joint optimization, the dynamic group-aware margin strategy based on reinforcement learning will identify the optimum margins for various age groups to mitigate the bias of imbalanced learning.}
	\label{fig_2}
\end{figure*}

In this section, we will elaborate on our proposed imbalanced learning framework GroupFace consisting of the Enhanced Multi-hop Attention Graph Convolutional Network and Dynamic Group-aware Margin Optimization. The overall pipeline is shown in Fig. \ref{fig_2}.

\subsection{Enhanced Multi-hop Attention Graph Convolutional Network}	
Aging changes between facial key points profoundly affect the accuracy of age estimation. However, common GCNs only focus on one-hop nodes to aggregate information from local domains, and simply deepening the network easily leads to over-smoothing issues. To improve the model for capturing the long dependencies between distant nodes, we introduce a multi-hop approach to expand the graph model receptive field.  

\begin{figure}[!t]
	\centering
	\includegraphics[width=0.48\textwidth]{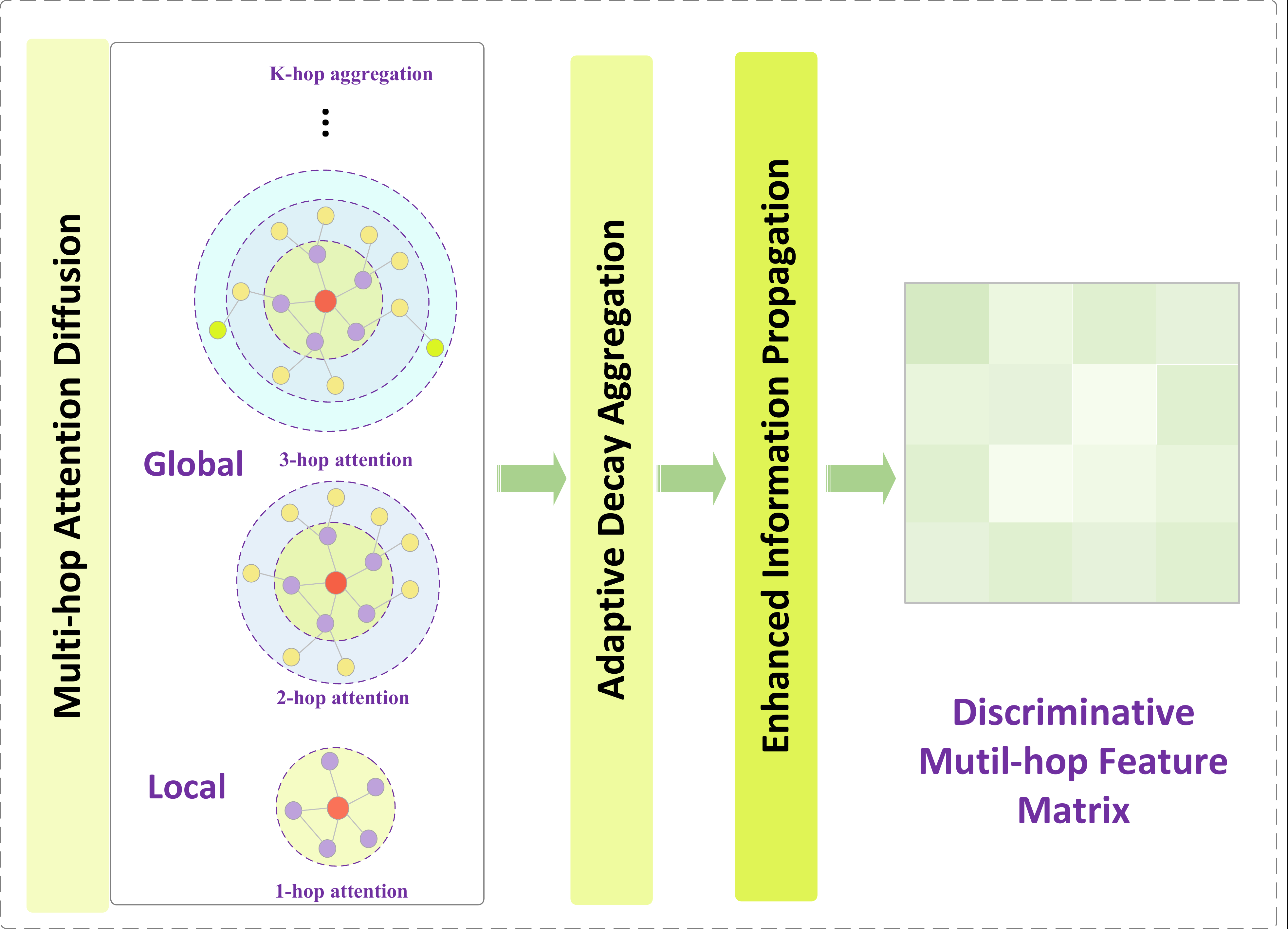}
	\caption{The illustration of \textcolor{black}{the main} designs of EMAGCN to capture discriminative features fusing global and local information.}
	\label{fig_3}
\end{figure}

Specifically, following \cite{wang2020multi}, we address some of the shortcomings of multi-hop networks with enhancements. The main design of our Enhanced Multi-hop Attention Graph Convolutional Network (EMAGCN) is shown in Fig. \ref{fig_3}. Since previous work aggregating different distance information with fixed weights tends to introduce higher-order noise and degrade the performance, EMAGCN employs an adaptive decay strategy to adaptively learn useful information.
We also randomly drop messages during message propagation to prevent over-fitting, and incorporate residual blocks in graph convolution to prevent over-smoothing. Finally, to address the problem of inefficient computation of dense matrices for higher-order graph models, we utilize a power iteration method to approximate the inverse matrix for acceleration.

\subsubsection{\textbf{Graph Construction}}
Formally, consider that an input face image of shape $H\times W\times 3$, we segment it into $N$ equal size patches and each patch is transformed into feature embedding $x_i\in\mathbb{R}^d$ \textcolor{black}{utilizing the convolution-based patch embedding methods \cite{zhang2024multi}, where $d$ denotes the feature dimensions. Regarding these patches representations $\mathcal{X}=\left\{ x_1,\,\,x_2,\,\,...\,\,,\,\,x_N \right\} $ as nodes $\mathcal{V}=\left\{ v_1,\,\,v_2,\,\,...\,\,,\,\,v_N \right\} $, we employ the K-Nearest algorithm \cite{han2022vision} to  calculate the neighbors $\mathcal{N}\left( v_i \right) $ of each node to build the edge.} Then we obtain a graph representation $\mathcal{G}=\left( \mathcal{V},\,\,\mathcal{E},\,\,\mathcal{A},\,\,\mathcal{M} \right)$, where $\mathcal{E}$ is the set of all edges between nodes, $\mathcal{A}$ denotes the adjacency matrix represented relation, $\mathcal{M}$ is the message matrix of the message-passing GNNs. The $A_{ij}$ is initialized to 1 when nodes $i$ to $j$ have edges, otherwise to 0, while $D=\sum_j{A_{ij}}=\left\{ d_1,\,\,...\,\,,\,\,d_N \right\} $ denotes the degree matrix, $d_i$ denotes the edge weight sum of node $v_i$.   
\textcolor{black}{The $\mathcal{M}=\left\{ m_1,\,\,...\,\,,\,\,m_N \right\} $ are 
	first initialized based on each node's own feature vector, and during each layer of message passing, updates the message matrix based on the features and edges of its neighboring nodes. Specifically, The message matrix propagate from node $i$ to node $j$ at $l$-th layer can be formulated as $M_{ij}^{\left( l \right)}=AGG_{j\in N\left( i \right)}\left( h_{i}^{\left( l \right)},h_{j}^{\left( l \right)},e_{ij} \right) $, where $h_{i}$ is the hidden representation of node $v_i$. }

\subsubsection{\textbf{Multi-hop Attention Diffusion}}
Similar to Graph Attention Network (GAT) \cite{velivckovic2017graph}, the first step is computing the attention values for all edges, the attention value of nodes $i$ to $j$ can be calculated as:
\begin{equation}
	\begin{split}
		\alpha _{ij}=\sigma \left( a_{\left( l \right)}^{T}\tan\text{h}\left( W_{i}^{\left( l \right)}h_{i}^{\left( l \right)}\parallel W_{j}^{\left( l \right)}h_{j}^{\left( l \right)} \right) \right) 
	\end{split}
	\label{equ1}
\end{equation}
where $h_{i}^{\left( l \right)}$ denotes the node $i$ embedding at $l$-th layer, and $h_{i}^{\left( 0 \right)}=x_i$. $a_{\left( l \right)}^{T}$, $W_{i}^{\left( l \right)}$ and $W_{j}^{\left( l \right)}$ are the learnable weights shared by $l$-th layer, $\sigma \left( \cdot \right) $ is the $LeakyReLU$ activation function and $\parallel $ denotes the concatenation operation. 

For all edges of $l$, we calculate 1-hop correlation by Eq. \ref{equ1}, then obtain an attention value matrix:
\begin{equation}
	\begin{split}
	S_{ij}=\left\{ \begin{array}{l}
		\,\,\,\,\alpha _{ij},\,\,\,\,if\,\,A_{ij}=1\\
		-\infty ,\,\,\,\,otherwise\\
	\end{array} \right. 
	\end{split}
	\label{equ2}
\end{equation}

We further apply $softmax$ operation on $S_{ij}$ to acquire the attention matrix at $l$-th layer:
\begin{equation}
	\begin{split}
	A_{ij}^{\left( l \right)}=sofmax \left( S_{ij} \right) 
	\end{split}
	\label{equ3}
\end{equation}
	
\subsubsection{\textbf{Adaptive Decay Aggregation}}
Subsequently, we utilize the graph diffusion method to compute the attention among nodes that lack a direct connection. \textcolor{black}{As shown in the upper two schematics in Fig. \ref{fig_4}, computing multi-hop attention allows creating attention shortcuts between nodes without explicit connection, utilizing the attention dependent on both their previous layer representation and the path relations between the nodes, thus effectively capturing long-distance interactions.} However, as the distance increases, the correlation between nodes becomes weaker and weaker, which tends to introduce useless information or noise. To overcome this weakness, we employ an adaptive decay strategy that separates semantic correlations of the network from different hops and assigns adaptive learnable decay weights, the power of the K-hop attention matrix is formulated as:
\begin{equation}
	\begin{split}
		\mathcal{A}=\sum_{k=0}^K{\bar{A}^k\odot \delta _k} 
	\end{split}
	\label{equ4} \tag{4}
\end{equation}
\textcolor{black}{where $\bar{A}^k$ is the power of attention matrix, such as $\bar{A}_{ij}^{k}$ represents the relational path number from node $i$ to $j$ of maximum length $k$, which can increase the receptive domain of attention. The $\delta _k=\left\{ \delta _1,\,\,...\,\,,\,\,\delta _k \right\} $ denotes the adaptive learnable attention decay factor adjusted by a $Sigmoid$ function $\delta _k=\frac{1}{1+e^{-\omega _k}}$ , where $\omega _k$ is the weight parameter learned by the model during the training process. Since the attention decay factor is adaptive rather than fixed, the learned weights are different for different hops and paths with different levels of importance, thus providing more flexibility in learning useful information while suppressing noisy information. Then the attention diffusion $AD\left( \cdot \right) $ to updated graph representation can be defined as $AD\left( \cdot \right) =\mathcal{A}H^{\left( l \right)} $.}

Furthermore, the attention diffusion equation for each individual head $i$ is computed distinctly as follows:

\begin{equation}
	\begin{split}
	\tilde{H}^{\left( l \right)}=MSA\left( \hat{H}^{\left( l \right)},\,\,\mathcal{G} \right) =\left( \parallel _{i=1}^{M}head_i \right) W_h
	\\
		head_i=AD\left( \mathcal{G},\,\,\tilde{H}^{\left( l \right)},\varTheta _i \right) ,\,\,\hat{H}^{\left( l \right)}=LN\left( H^{\left( l \right)} \right)
	\end{split}
	\label{equ5}
\end{equation}
where $MSA\left( \cdot \right) $ is the multi-head self attention, $AD\left( \cdot \right) $ denotes the attention diffusion, $LN\left( \cdot \right)$ is the layer normalization for stabilizing the computation procedure, $W_h$ is a parameter matrix and $\varTheta _i$ is the parameter of the $i$-th head.

With attention diffusion, the receptive domain of the graph model is enlarged and global and local information are effectively captured, while the decay strategy reduces redundant information from distant nodes.

\subsubsection{\textbf{Enhanced Information Propagation}}
During the node information propagation process, we randomly drop some messages \cite{fang2023dropmessage} for enhancement to maintain the diversity of topological information as well as to prevent the training of long-tailed samples from over-fitting. Unlike the previous drop methods, DropMessage directly drops the message matrix $M$. By dropping the message matrix with drop ratio $\varrho $, the process can be elaborated as:
\begin{equation}
	\begin{split}
	\tilde{M}_{ij}=\frac{1}{1-\varrho}\epsilon _{ij}M_{ij}
	\end{split}
	\label{equ6}
\end{equation}
where $\epsilon _{ij}\sim Bernoulli\left( 1-\varrho \right) $ denotes an independent mask determining whether a reservation will be made. Thus node representation is updated based on node feature information and messages from multi-hop neighbors, which can be represented as: 
\begin{equation}
	\begin{split}
	h_{i}^{\left( l+1 \right)}=h_{i}^{\left( l \right)}+\tilde{M}_i
	\end{split}
	\label{equ7}
\end{equation}

Moreover, considering the significant increase in the computation of dense matrices generated by the multi-hop graph model, we introduce a power iteration method to estimate the inverse matrix, thereby attaining linear complexity accelerating training, and reducing computational overhead. The specific computational procedure can be calculated as:
 
\begin{flalign}
	\begin{split}
			H^{\left( n \right)}=Q^TH^{\left( n-1 \right)}+H^{\left( 1 \right)} \\
			=Q^nX+Q^{n-1}X+...+Q^2X+Q^TX
		\label{equ8}
		\end{split}
\end{flalign}
where $Q=\tilde{A}\tilde{D}^{-1}$ is a row normalization instead of standard Laplacian smoothing normalization and $n$ is the iterations. When $n$ tends to infinity, the matrix hierarchy $\left( I-Q \right) ^{-1}$ converges to $I+M+M^2+...+M^{n-1}+M^n$. 

\begin{figure}[!t]
	\centering
	\includegraphics[width=0.48\textwidth]{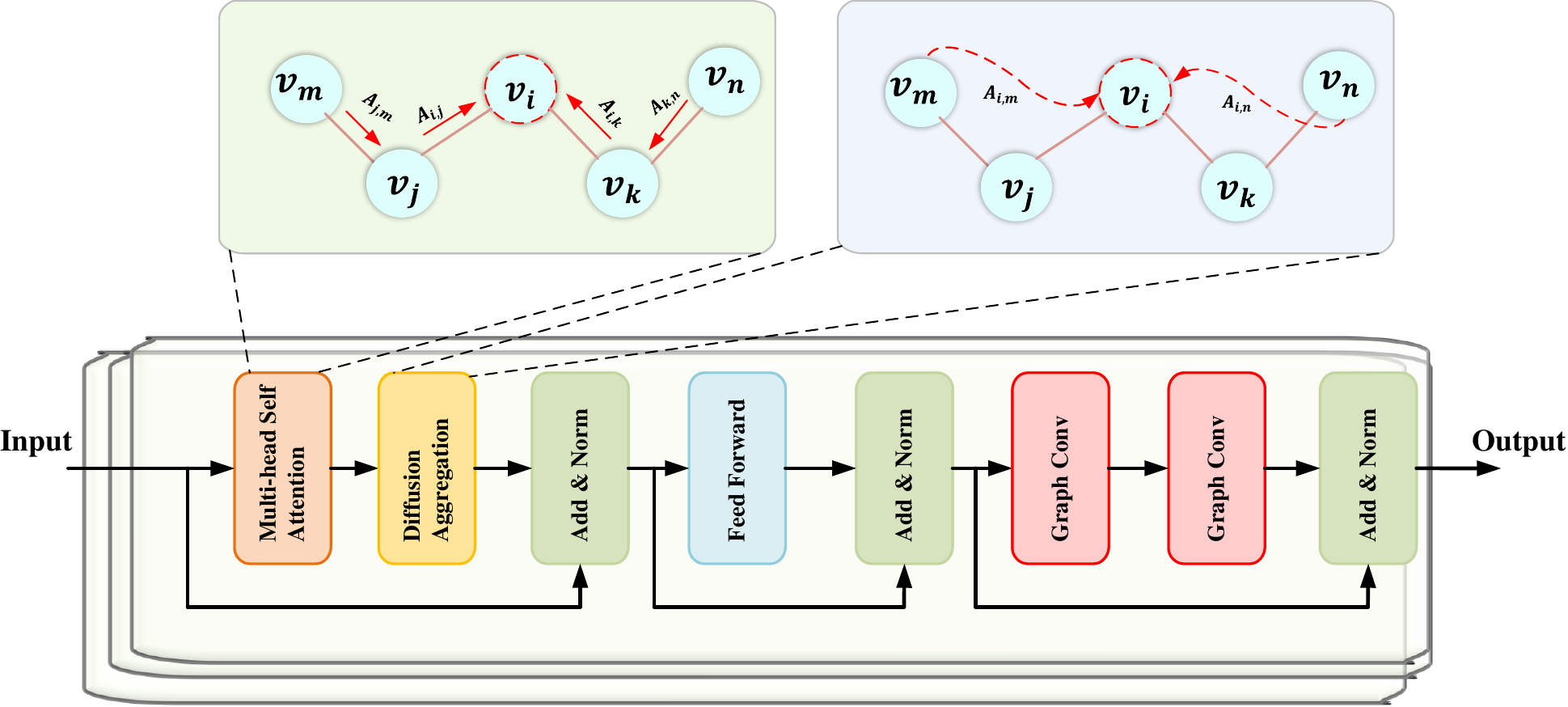}
	\caption{The illustration of EMAGCN blocks. In the schematic above, the relations between multi-hops are obtained by attention diffusion.}
	\label{fig_4}
\end{figure}

Finally, EMAGCN \textcolor{black}{propagates} all information into the face features through the process of graph convolution, as shown in Fig. \ref{fig_4}, which \textcolor{black}{contains} a fully connected feed-forward sublayer, add layer normalization and residual connections to achieve a more expressive information propagation process:

\begin{flalign}
	\begin{split}
	&\tilde{H}^{\left( l+1 \right)}=\tilde{H}^{\left( l \right)}+H^{\left( 0 \right)},
	\\
	&H^{\left( l+1 \right)}=\sigma \left( LN\left( \tilde{H}^{\left( l+1 \right)} \right) W_{1}^{\left( l \right)} \right) W_{2}^{\left( l \right)}+\tilde{H}^{\left( l+1 \right)}
	\label{equ9}
	\end{split}
\end{flalign}
where  $\sigma \left( \cdot \right) $ denotes the $LeakyReLU$ activation function, $LN\left( \cdot \right) $ is the layer normalization and $W_{1}^{\left( l \right)} , W_{2}^{\left( l \right)}$ are the different trainable weight. 

\subsection{Dynamic Group-aware Margin Loss}	
Large margin loss functions based on softmax are often used to train feature extractors to make learned features more discriminative. A unified form is defined as:
\begin{equation}
	\begin{split}
	\mathcal{L}=-\frac{1}{N}\sum_{i=1}^N{\log}\frac{e^{f\left( \theta _{y_i},m \right)}}{e^{f\left( \theta _{y_i},m \right)}+\sum\limits_{j\ne y_i}^n{e^{s\cos \theta _j}}}
	\end{split}
	\label{equ10}
\end{equation}
where $y_i$ is the label index, $m$ denotes the margin, and $\theta _{y_i}$ denotes the angle between the weight and feature vector of the $j$-th classifier.

However, most margins \cite{wang2018cosface, deng2019arcface } use fixed values, which are not flexible enough for real-life complex classification and have a large bias for category imbalance, so many studies \cite{boutros2022elasticface, xu2024x2} have begun to explore adaptive margins. 

\begin{figure}[!t]
	\centering
	\includegraphics[width=0.48\textwidth]{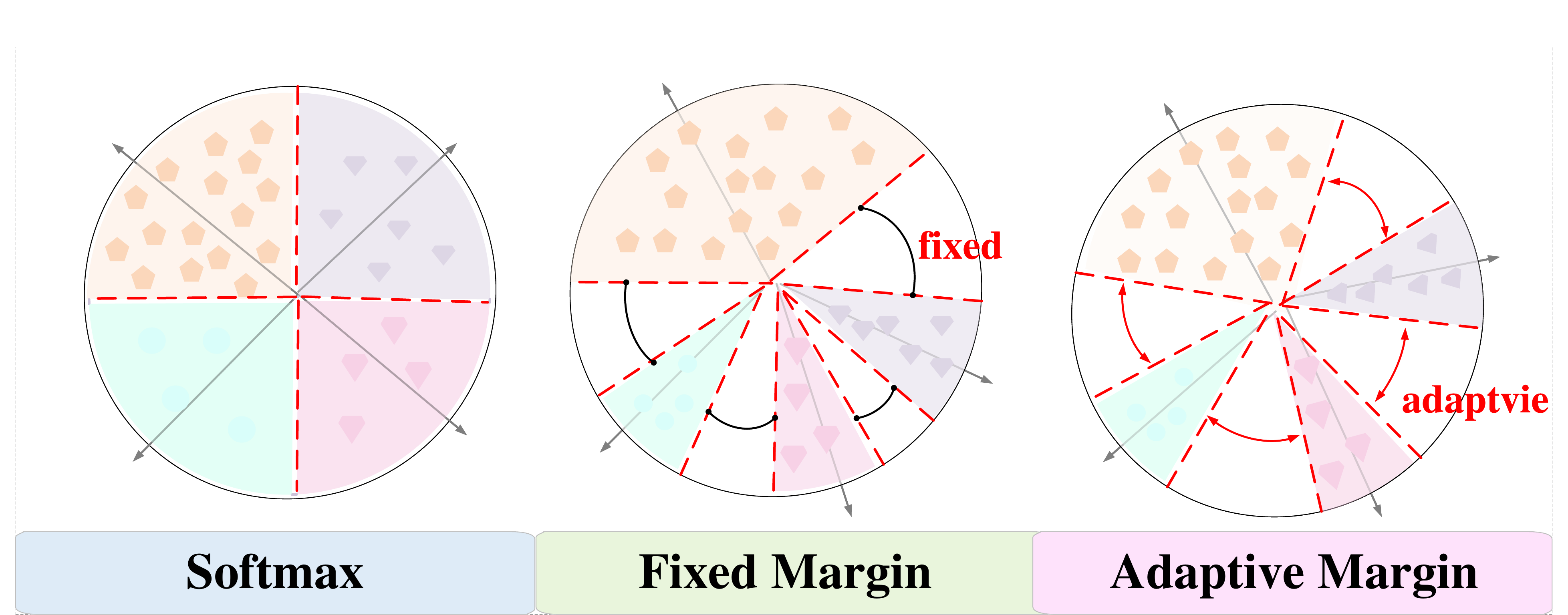}
	\caption{The illustration of different margin losses. GroupFace employs the dynamic group-aware margin loss, which can adaptively provide suitable and unbiased margins for different age groups.}
	\label{fig_5}
\end{figure}

Considering the poor generalization ability of long-tailed groups, we introduce the idea of dynamic adaptive into imbalanced age estimation to improve the large margin loss, which can be shown in Fig. \ref{fig_5}. Our goal is to explore the adaptive margins between different classes, which can be guided by subsequent reinforcement learning, and the angular margins can be adjusted by the loss function automatically. By balancing the margins between different classes, it is realized that the majority class will not converge with too large a gap, while the minority class will converge to the majority class with a smaller gap. Following the \cite{xu2024x2}, we design a dynamic group-aware margin loss to balance the additional margins between different classes, which can be formulated as:
\begin{equation}
	\begin{split}
		\mathcal{L}_{DGM}=-\frac{1}{N}\sum_{i=1}^N{\log}\frac{e^{a_i\left( t \right) \left( \theta _j-h_i\left( t \right) \right) ^2+k_i\left( t \right)}}{e^{a_i\left( t \right) \left( \theta _j-h_i\left( t \right) \right) ^2+k_i\left( t \right)}+\sum\limits_{j\ne y_i}^n{e^{s\cos \theta _j}}} 
	\end{split}
	\label{equ14}
\end{equation}
where $a_i\left( t \right)$, $h_i\left( t \right)$ and $k_i\left( t \right)$ are the adaptive margin parameters associated with different groups, which can be determined in the reinforcement learning training stage.

By converting the cosine function to a quadratic function, model over-fitting can be avoided and it helps to reduce the computation overhead. Three learnable parameters make the margins more adaptive between different groups, thus enhancing the inter-class discrimination and intra-class compactness of age estimation features.

 \subsection{RL-based Dynamic Group-aware Margin Optimization}	
Inspired by \cite{liu2019fair, wang2020mitigating}, we conceptualize the problem of identifying suitable adaptive margins within the framework of a Markov Decision Process (MDP).  With state $s_t$ as input, the Q-value $Q\left( s_t, a \right)$ is estimated by the deep Q-learning network (DQN), where the agent will be trained to use different actions $a_t$ to adapt the margins for each state. In turn, the environment will give the action a reward $r\left( s_t,a_t \right)$, and then update the state by $s_{t+1}$. The training objective of deep Q-learning is to seek the optimal function $Q^*$, which indicates that the agent can obtain the highest cumulative reward value by adjusting the margins of each iteration with policy $\pi$.

\subsubsection{\textbf{State}}	
The dynamic group-aware margin is adaptively adjusted to the number of images in different groups, inter-class, and intra-class distance, which is defined by the triple $\left\{ G,\,\,D_{inter},\,\,M \right\}$. We divide the age groups into four categories $G=\left\{ 0,1,2,3 \right\} $ where $Children$ (group 0), $Teenager$ (group 1), $Adult$ (group 2) and $Senior$ (group 3). $M$ is equivalent to the dynamic group-aware margin. 
\textcolor{black}{
Since the number of the $Adult$ age group in most age datasets is the largest, we keep the margin of $Adult$ as the anchor and achieve imbalanced learning by adjusting the angular skewness of the long-tailed age groups with respect to the head class $Adult$.
}
$D_{inter}$ denotes the deviation of the inter-class distance between age group $i$ of the long-tailed classes and the head class $Adult$, which can be expressed as:
\begin{equation}
	\begin{split}
		D_{inter}=|d_{inter}^{i}-d_{inter}^{2}|
		\\
		d_{inter}^{i}=\frac{1}{N_i}\sum_{i=1}^{N_i}{\max_{k=1:N_i}\cos \left( x_k,x_i \right)} 
	\end{split}
	\label{equ15}
\end{equation}
where $d_{inter}^{i}$ is the inter-class distance of $i$-th age group, $d_{inter}^{2}$ denotes the inter-class distance of the head class $Adult$. Moreover, $\cos \left( \cdot, \cdot \right) $ represents the cosine distance function, $N_i$ is the $i$-th age group's number and $x_i$ denotes the feature center of $i$-th age group.

Assuming that different age groups have different requirements for margins, these requirements may vary depending on their $D_{inter}$. When the bias is large, long-tailed groups may require larger margins to improve their generalization ability. In addition, to make the space of states discrete, we let $D_{inter}$ and $M$ identified to discrete space $\mathcal{D}$ and $\mathcal{M}$, where $\mathcal{D}=\left\{ d_1,d_2,...,d_{nD} \right\} $ and $\mathcal{M}=\left\{ m_1,m_2,...,m_{nM} \right\} $.

\subsubsection{\textbf{Action}}
Depending on the different state, there are three types of action spaces $\mathcal{A}=\left\{ -1, O, +1 \right\} $, where $O$ means keep the same, $-1$ means shrink to a constant value $\kappa$ and $+1$ means expand to a constant value $\kappa$.

We train the agent to make better decisions while taking action $a$ obeying the cumulative reward $\pi ^*\left( s, a \right) =arg\max _aQ\left( s, a \right) $. For instance, if at time step $t$ the agent opts to execute action $-1$ based on the $Q$ value and the state $s_t=\left\{ 3,\,\,d_1,\,\,m_1 \right\}$, the margin of the Senior will be updated to $m_2=m_1-\kappa$. 

\subsubsection{\textbf{Reward}}
The reward function $r\left( s_t,a_t \right) $ is set to motivate the agent to take a better action $a_t$ at state $s_t$. We expect that the long-tailed classes have close generalization ability of the head class $Adult$, then the age estimation biases of the different groups are balanced, and so we utilize the deviation of the inter-class and intra-class distances to form the rewards. The deviation of intra-class distances between age group $i$ of the long-tailed classes and the head class $Adult$ can be expressed as:
\begin{equation}
	\begin{split}
	D_{intra}=|d_{intra}^{i}-d_{intra}^{2}|
	\\
	d_{intra}^{i}=\frac{1}{N_i}\sum_{i=1}^{N_i}{\cos \left( x_k,x_i \right)} 
	\end{split}
	\label{equ16}
\end{equation}
where $d_{intra}^{i}$ denotes the intra-class distance of $i$-th age group , $d_{intra}^{2}$ denotes the intra-class distance of the head class $Adult$. $N_i$ is the number of $i$-th age group, and $x_i$ denotes the feature center of  $i$-th age group. Combining the deviation of the inter-class and intra-class distances together, the reward to adjust the margin can be formulated as:
\begin{equation}
	\begin{split}
		r\left( s_t,a_t \right) =\mathcal{R}_{t+1}-\mathcal{R}_t
		\\
		\mathcal{R}=-\left( D_{intra}+D_{inter} \right) 
	\end{split}
	\label{equ17}
\end{equation}

\subsubsection{\textbf{Deep Q-Learning}}
It is employed to \textcolor{black}{seeks} the optimal function $Q^*$ to guide the agent acquiring the highest cumulative reward value. During the training process, we iteratively change the Q-function to update the model by minimizing the loss:
\begin{equation}
	\begin{split}
		\mathcal{L}_{DQN}=\mathbb{E}_{s_t,a_t}\lVert y_t-Q\left( s_t,a_t \right) \rVert ^2
		\\
   y_t=\mathbb{E}_{\pi}\left( r_t+\gamma \max Q\left( s_{t+1},a_{t+1}|s_t,a_t \right) \right) 
	\end{split}
	\label{equ18}
\end{equation}
where $y_t$ is the target $Q$ value, $\gamma$ is the discount factor, $y_t-Q\left( s_{t,}a_t \right)$ denotes the deviation error at $t$-th step and a future reward $\gamma \max Q\left( s_{t+1},a_t|s_t,a_t \right)$.

\subsubsection{\textbf{Training Network}}
The samples are used as inputs to train the DQN, and the agent dynamically adapts the margin through actions for different groups. After traversing all the states through continuous iterative updating, the optimal group-aware margin policy will be obtained.

Then the group-aware margin policy will guide and optimize our overall age estimation network, including robust extraction of face features and accurate classification of age groups. Our age estimation method starts with age classification at smaller intervals through the four age groups, and the classification results are further regressed to obtain predicted values.

In the multi-classification process, for the $i$-th element output $z_i$, softmax is employed for age grouping to $g$ categories with $G_i=Soft\max \left( z_i \right) =\frac{e^{z_i}}{\sum\limits_{j=1}^g{e}^{z_j}}
$. After the softmax operation, the predicted age with label $y_i$ is calculated by the expectation $\hat{y}_i=E=\sum_{i=0}^{|g-1|}{y_i.G_i}$. The joint two-stage estimation loss consists of cross-entropy loss and average absolute loss, which is formulated as:
\begin{equation}
	\begin{split}
	L_{GroupFace}=\lambda L_{CE}+\left( 1-\lambda \right) L_{MAE} 
	\end{split}
	\label{equ19}
\end{equation}

\section{Experiment}
In this section, we present our experiments in detail, which will be organized around the following questions: 

\textbf{RQ1:} \emph{How effective and robust is our method under general age estimation?}

\textbf{RQ2:} \emph{Can our method achieve balanced generalization performance in long-tailed age estimation?}

\textbf{RQ3:} \emph{How do the key designs and hyperparameters in the architecture affect the performance?}

\subsection{Datasets}
\subsubsection{\textbf{MORPH-II}}
The dataset \cite{ricanek2006morph} is a widely used age estimation benchmark comprising 55,134 facial images from 13,617 individuals, ages spanning from 16 to 77. It also provides demographic details like gender, race, and glasses usage. In our experiments, we used two protocols in our evaluation. \textbf{Setting I}  \cite{gao2018age}: The entire dataset is randomly 
segmented into two distinct sections, with 80$\%$ allocated for training and 20$\%$ reserved for testing. \textbf{Setting II} \cite{tan2017efficient}: Based on a subset of 5493 facial images of Caucasian ethnicity, the subset was randomly segmented into two sections: 80$\%$ and 20$\%$ for training and testing.

\subsubsection{\textbf{UTK-Face}}
The dataset \cite{zhang2017age} is a large-scale, long age-span (from 0 to 116 years) face dataset. Collected in an unconstrained setting, it includes over 20,000 images labeled with gender, age, and ethnicity, capturing a broad spectrum of variations. In our work, we randomly select 80$\%$ and 20$\%$ for training and testing.

\subsubsection{\textbf{Chalearn LAP 2015}}
The dataset \cite{escalera2015chalearn} is a competitive dataset released at the ChaLearn LAP Challenge 2015 and contains 4699 face images with annotations averaged from at least 10 users. It is divided into training (2476 images), validation (1136 images), and testing (1079 images) subsets.

\subsubsection{\textbf{CACD}}
The dataset \cite{chen2015face} is a vast dataset from 2,000 celebrities with 163,446 images aged 14 to 62. Due to estimated age annotations, it has more noise. In our experiments, we cleansed the noise and utilized 1800 celebrities for training, and 80 and 120 for validation and testing.

\subsubsection{\textbf{MIVIA}}
The dataset \cite{greco2022effective}, derived from the CAIP competition and sourced from VGGFace2, contains 575,073 images labeled with ages from 1 to 81 using ``knowledge distillation." With the test set pending release, MIVIA was split into 458,752 and 114,688 images for training and validation.

\begin{table*}[!t]
	\caption{The summary of widely-used age estimation datasets. Some detailed information is from \cite{agbo2021deep}.}
	\renewcommand\arraystretch{1.4}
    \tabcolsep=0.031\linewidth
	\centering
	\begin{tabular}{c|cccccc}
		\hline
		\textbf{Dataset} & \textbf{\#Images} & \textbf{\#Subjects} & \textbf{Age range} & \textbf{Label type} & \textbf{In-the-wild} & \textbf{Class imbalance} \\ \hline
		MORPH II\cite{ricanek2006morph}         & 55,134            & 13,618              & 16-77              & Real age            & No                   & Yes                      \\
		UTK-Face\cite{zhang2017age}          & 23,708            & -                   & 0-116              & \textcolor{black}{Apparent} age       & Yes                  & Yes                      \\
		CLAP15\cite{escalera2015chalearn}           & 4,691             & -                   & 3-85               & \textcolor{black}{Apparent} age       & Yes                  & Yes                      \\
		CACD\cite{chen2015face}             & 163,336           & 2000                & 14-62              & \textcolor{black}{Apparent} age       & Yes                  & Yes                      \\
		MIVIA\cite{greco2022effective}            & 575,073           & -                   & 1-81               & \textcolor{black}{Apparent} age       & Yes                  & Yes                      \\ \hline
	\end{tabular}
	\label{table1}	                   
\end{table*}

\subsection{Evaluation Metrics}
\subsubsection{\textcolor{black}{\textbf{MAE (Mean Absolute Error)}}}
It is defined as the average distance between the actual and predicted age, which is formulated as:
\begin{equation}
	\begin{split}
	MAE=\frac{1}{N}\sum_{i=1}^N{|}y_i-\hat{y}_i|
	\end{split}
	\label{equ20}
\end{equation}
where $y_i$ and $\hat{y}_i$ are the true and predicted age values of the $i$-th sample, respectively, and $N$ is the amount of test images. 

\subsubsection{\textbf{$\epsilon$-error (Normal Score)}}
The age in the CLAP-2015 dataset is labeled as the mean of different people, and the true age in the data contains two attributes, mean and variance. Thus considering these factors can be a more accurate measure, which is calculated as:
\begin{equation}
	\begin{split}
	\epsilon =1-\sum_{i=1}^N{e}xp\left( -\frac{\left( y_i-\hat{y}_i \right) ^2}{2\sigma _i^2} \right) 
	\end{split}
	\label{equ21}
\end{equation}
where $y_i$ and $\hat{y}_i$ denote the true and predicted age values of the $i$-th sample, respectively, $N$ is the amount of test images, and $\sigma _i^2$ is the labeled standard deviation. The smaller the $\epsilon$-error, the more accurate the age estimate.

\subsubsection{\textcolor{black}{\textbf{AAR (Age Accuracy and Regularity)}}}
To further evaluate the performance of the model under imbalanced or long-tailed distributions, we introduce the protocol of \cite{greco2022effective,bao2021lae}. This metric can assume values between 0 and 10, weighted with 70$\%$ MAE (accuracy) and 30$\%$ $\sigma$ (regularity) contributions, which can be formulated as:
\begin{flalign}
	\begin{split}
	&AAR=\,\,\max \left( 0;7-MAE \right) +\,\,\max \left( 0;3-\sigma \right), 
	\\
		&\sigma =\sqrt{\frac{\sum\limits_{j=1}^n{\left( MAE^j-MAE \right) ^2}}{n}} .
	\end{split}
	\label{equ22}
\end{flalign}
where $n$ is the number of age groups, $MAE$ is the mean absolute error of the entire test set, $\sigma$ denotes the standard deviation of different age groups, and $MAE^j$ is the MAE computed for the sample whose actual age is in the $j$-th age group.

\subsection{Implementation Details}
We first utilize the face landmark algorithm MTCNN \cite{zhang2016joint} to detect and align each face image, and then crop them to 224 * 224. During the training process, the samples are augmented with translations, color dithering,  and random rotations. The starting learning rate was established at 0.0001 for the entirety of the experiments and attenuated using a cosine annealing strategy. We used the Adam optimizer \cite{kingma2014adam} with parameters for weight decay and momentum configured to 0.0005 and 0.9, respectively. Each model underwent training on NVIDIA RTX 3090 GPUs using PyTorch.

\subsection{Comparisons With The State-of-the-Art Methods (\textbf{RQ1})}
To validate the efficacy of our proposed GroupFace, \textcolor{black}{detailed} experiments \textcolor{black}{are} conducted on three face image datasets. For the characteristics of different datasets, we utilize appropriate experimental settings and evaluation criteria to compare them with state-of-the-art methods (SOTAs).

\subsubsection{\textbf{Comparisons on Morph II}}
On the most popular restricted datasets, TABLE \ref{table2} lists the avenue year, backbone network, and number of parameters for the SOTAs. 
\textcolor{black}{
Our method achieves the MAEs of 2.09 and 1.86 (with IMDB-WIKI dataset pre-trained weights) under Setting I, which outperforms PML\cite{deng2021pml}, DCT\cite{bao2022divergence}, and MSL\cite{wang2023exploiting} that also consider imbalanced learning. It is inferior to GLAE\cite{bao2023general} pre-trained on extra datasets using MS-CELEB-1M, TAA-GCN\cite{korban2023taa}, MetaAge\cite{li2022metaage}, and MWR\cite{shin2022moving} without using extra datasets but is able to achieve similar performance to the SOTAs with the IMDB-WIKI dataset weights. It is worth noting that compared to the commonly used age estimation pre-training dataset IMDB-WIKI, MS-CELEB-1M is far superior to it in terms of the number and quality of images. So a high-quality pre-training dataset and three excellent network designs enabled GLAE\cite{bao2023general} to achieve a performance that far exceeded the best of others. Under Setting II, our method achieves the second-best MAE 2.01 with the IMDB-WIKI dataset weights, which approaches the  GLAE. In addition, our network has 8.6 M parameters, which is smaller than these methods with the highest prediction accuracy.
}

\begin{table*}[!t]
	\caption{Comparisons with the state-of-the-art methods on Morph II. (\textcolor{black}{* indicates used the IMDB-WIKI dataset for pre-training,$^{\dagger}$ indicates used the MS-CELEB-1M dataset for pre-training, and ‘$\downarrow$’ indicates the smaller is better.})}
	\renewcommand\arraystretch{1.4}
    \tabcolsep=0.039\linewidth
	\centering
	\begin{tabular}{c|ccccc}
		\hline
		\multirow{2}{*}{\textbf{Method}} &
		\multirow{2}{*}{\textbf{Venue Year}} &
		\multirow{2}{*}{\textbf{Backbone Network}} &
		\multicolumn{2}{c}{\textbf{MAE$\downarrow$}} &
		\multirow{2}{*}{\textbf{Param.$\downarrow$}} \\ \cline{4-5}
		&            &                  & \textbf{Setting I} & \textbf{Setting II} &       \\ \hline
		DEX \cite{rothe2015dex}          & IJCV 2016  & VGG-16           & -                  & 3.15/2.68*          & 138M  \\
		Ranking-CNN \cite{chen2017using} & CVPR 2017  & Binary CNNs      & 2.96*              & -                   & 500M  \\
		DLDLF \cite{shen2017label}       & NIPS 2017  & VGG-16           & 2.24               & -                   & 138M  \\
		MV \cite{pan2018mean}            & CVPR 2018  & VGG-16           & 2.79 /2.16*        & -                   & 138M  \\
		SSR-NET \cite{yang2018ssr}       & IJCAI 2018 & SSR-NET          & 3.16*              & -                   & 40.9K \\
		C3AE \cite{zhang2019c3ae}        & CVPR 2019  & C3AE             & 2.78*              & 2.95*               & \textbf{39.7K} \\
		BridgeNet\cite{li2019bridgenet}  & CVPR 2019  & VGG-16           & 2.38*              & 2.35*               & 138M  \\
		PML \cite{deng2021pml}           & CVPR 2021  & ResNet-34        & 2.15               & 2.31                & 21M   \\
		MWR \cite{shin2022moving}        & CVPR 2022  & VGG-16            & 2.00*              & 2.13*              & 138M  \\
		MetaAge \cite{li2022metaage}         & TIP 2022  & VGG-16            & 1.81*              & 2.23*               & 138M  \\
		DAA \cite{chen2023daa}           & CVPR 2023  & ResNet-18        & 2.25/2.06*         & -                   & 11M   \\
		TAA-GCN \cite{korban2023taa}           & PR 2023  & TAA-GCN        & 1.69        & -                  & -   \\
		DCT \cite{bao2022divergence}          & TIFS 2023  & ResNet-50        & 2.28/2.17*         & -                  & 23M   \\
		MSL \cite{wang2023exploiting}           & TIFS 2023  & ResNet-34        & 2.10         & 2.03                 & 21M   \\
		\color{black} GLAE \cite{bao2023general}
		& \color{black} TIP 2023  & \color{black} ResNet-50        & \color{black} \textbf{1.14$^{\dagger}$}         & \color{black} \textbf{2.00$^{\dagger}$}                 & \color{black} 23M   \\ 
		\cellcolor{gray!15}\textbf{GroupFace (Ours)} &
		\cellcolor{gray!15}\textbf{-} &
		\cellcolor{gray!15}\textbf{EMAGCN} &
		\cellcolor{gray!15}2.09/1.86* &
		\cellcolor{gray!15}2.27/2.01* &
		\cellcolor{gray!15}8.6M \\ \hline
	\end{tabular}
	\label{table2}	                   
\end{table*}

\subsubsection{\textbf{Comparisons on UTK-Face}}
We test the performance of GroupFace on an unconstrained large-scale dataset spanning ages 0 to 116. TABLE \ref{table3} shows that our GroupFace achieves the second-best MAE of 4.32 and performs far better than previous methods with a smaller number of parameters of 8.6M. Note that MSL\cite{wang2023exploiting} uses a deeper ResNet-34, and MWR \cite{shin2022moving} uses a larger VGG-16, both of which effectively reduce the MAE. Andrey Savchenko \cite{savchenko2019efficient} uses a more compact network MobileNet-v2, which has the lowest number of parameters but lacks in performance.

\begin{table}[!t]
	\caption{Comparisons with the state-of-the-art methods on UTK-Face. (\textcolor{black}{* indicates used the IMDB-WIKI dataset for pre-training, and ‘$\downarrow$’ indicates the smaller is better.})}
	\renewcommand\arraystretch{1.4}
    \tabcolsep=0.038\linewidth
	\centering
	\begin{tabular}{c|ccc}
		\hline
		\textbf{Method}                                       & \textbf{Backbone} & \textbf{MAE$\downarrow$}  & \textbf{Param.$\downarrow$} \\ \hline
		Andrey Savchenko  \cite{savchenko2019efficient}      & MobileNet-v2      & 5.44*          & \textbf{3.4M}   \\
		CORAL \cite{cao2020rank}                           & ResNet-50         & 5.47*          & 25.6M           \\
		DCDL \cite{sun2021deep}                              & VGG-16             & 4.48*          & 138M            \\
		MWR \cite{shin2022moving}                              & VGG-16             & 4.37          & 138M            \\
		MSL \cite{wang2023exploiting}             & ResNet-34        & \textbf{4.31*}                  & 21M   \\
		\cellcolor{gray!15}\textbf{GroupFace (Ours)}                              & \cellcolor{gray!15}\textbf{EMAGCN}    & \cellcolor{gray!15}4.32\color{black}*& \cellcolor{gray!15}8.6M            \\ \hline
	\end{tabular}
	\label{table3}
\end{table}

\subsubsection{\textbf{Comparisons on ChaLearn LAP 2015}}
We further compare our method with the SOTAs on the unrestricted competitive dataset CLAP 2015. TABLE \ref{table4} shows our results, where we achieve the third best MAE of 2.91 after GLAE\cite{bao2023general} and DCT \cite{bao2022divergence}, and the best $\epsilon$-error of 0.239, while the parameter count of 8.6M is the smallest among these methods. GLAE designed Adaptive Routing (AR) to select suitable classifier for improving the long-tailed recognition while maintaining the head class. \textcolor{black}{However, our method achieves similar performance with the help of the IMDB-WIKI dataset while GLAE with the help of MS-CELEB-1M, indicating that GroupFace is also effective in handling samples with large variance.}

\begin{table}[!t]
	\caption{Comparisons with the state-of-the-art methods on ChaLearn LAP 2015. (\textcolor{black}{* indicates used the IMDB-WIKI dataset for pre-training, $^{\dagger}$ indicates used the MS-CELEB-1M dataset for pre-training, and ‘$\downarrow$’ indicates the smaller is better.})}
	\renewcommand\arraystretch{1.4}
    \tabcolsep=0.05\linewidth
	\centering
\begin{tabular}{c|ccc}
	\hline
	\textbf{Method}                & \textbf{MAE$\downarrow$}   & \textbf{$\epsilon$-error$\downarrow$} & \textbf{Param.$\downarrow$} \\ \hline
	DEX \cite{rothe2015dex}         & 3.25*          & 0.282            & 138M            \\
	DHAA \cite{tan2019deeply}       & 3.05*          & 0.265            & 100M            \\
	BridgeNet \cite{li2019bridgenet} & 2.98*          & 0.267            & 138M            \\
	DLDL \cite{gao2017deep}         & 3.51           & 0.310            & 138M            \\
	AGEn \cite{tan2017efficient}    & 3.21           & 0.280            & 138M            \\
	PML \cite{deng2021pml}          & 2.91*          & 0.243            & 21M             \\
	MWR \cite{shin2022moving}        & 2.95*           & 0.262            & 138M            \\
	DCT \cite{bao2022divergence}    & 2.87$^{\dagger}$ & 0.242            & \color{black}11M             \\
	GLAE\cite{bao2023general}    & \textbf{2.852$^{\dagger}$} & 0.242            & \color{black}11M             \\
	\cellcolor{gray!15}\textbf{GroupFace (Ours)}       & \cellcolor{gray!15}
	2.91\color{black}*           & \cellcolor{gray!15}\textbf{0.239}   & \cellcolor{gray!15}\textbf{8.6M}   \\ \hline
\end{tabular}
	\label{table4}
\end{table}

\subsubsection{\textbf{Comparisons on CACD}}
We also compare our GroupFace with the SOTAs on the large dataset CACD, which originates from web crawling with a lot of noisy data and large variations in face background and illumination. As shown in TABLE \ref{table5}, our framework achieves an optimal performance of 4.07 MAE with IMDB-WIKI dataset pre-trained weights. Compared with GLAE\cite{bao2023general}, MSL\cite{wang2023exploiting} and DCT \cite{bao2022divergence}, the amount of parameters is nearly halved thanks to the flexibility of the graph model, although our model only reduces 0.02 MAE. Clearly, the results show that our GroupFace is capable of unconstrained age estimation and robust feature extraction in samples containing more noise.

\begin{table}[!t]
	\caption{Comparisons with the state-of-the-art methods on CACD. (\textcolor{black}{* indicates used the IMDB-WIKI dataset for pre-training, $^{\dagger}$ indicates used the MS-CELEB-1M dataset for pre-training, and ‘$\downarrow$’ indicates the smaller is better.})}
	\renewcommand\arraystretch{1.4}
    \tabcolsep=0.09\linewidth
	\centering
    \begin{tabular}{c|cc}
    	\hline
    	\textbf{Method}             & \textbf{MAE$\downarrow$}  & \textbf{Param.$\downarrow$} \\ \hline
    	DEX \cite{rothe2015dex}     & 4.79*         & 138M            \\
    	DHAA \cite{tan2019deeply}    & 4.35*         & 100M            \\
    	CR-MTk \cite{liu2020facial} & 4.48*         & 67M             \\
    	MWR \cite{shin2022moving}    & 4.41          & 138M            \\
    	DCT \cite{bao2022divergence} & 4.09$^{\dagger}$         & 23M             \\
    	TAA-GCN \cite{korban2023taa} & 4.09         & -             \\
    	MSL \cite{wang2023exploiting} &  4.126*                 & 21M   \\
    	GLAE \cite{bao2023general} &  4.09$^{\dagger}$                 & 23M   \\
    	\cellcolor{gray!15}\textbf{GroupFace (Ours)}    & \cellcolor{gray!15}\textbf{4.07}\color{black}\textbf{*} & \cellcolor{gray!15}\textbf{8.6M}   \\ \hline
    \end{tabular}
	\label{table5}
\end{table}

Our GroupFace strikes a great improvement which probably owing to: i) GroupFace effectively integrates Enhanced Multi-hop Graph Convolutional Networks (EMAGCN) and reinforcement learning-based group-aware margin strategy, focusing on both discriminative feature extraction for different groups and imbalanced learning for the long-tailed class. ii) The graph model is more flexible in dealing with complex irregular objects that effectively model aging changes in the face, and reduces the learning of excessive redundant information thus making the model more compact.

\begin{table*}[!t]
	\caption{The long-tailed Generalization Analysis on three long-range and imbalanced age estimation datasets. \textcolor{black}{(‘↓’ indicates the smaller is better, while ‘↑’ indicates the larger the better.)}}
	\renewcommand\arraystretch{1.4}
	\setlength\tabcolsep{1.72mm}
	\centering
	\begin{tabular}{c|ccccccc|ccccccc}
		\hline
		\multirow{3}{*}{\textbf{Method}} & \multicolumn{7}{c|}{\textbf{MORPH II (Setting I)}}                                                                                 & \multicolumn{7}{c}{\textbf{MORPH II (Setting II)}}                                                                                 \\ \cline{2-15} 
		& \multicolumn{4}{c|}{\textbf{Group MAE}}                                            & \multicolumn{3}{c|}{\textbf{Overall}}         & \multicolumn{4}{c|}{\textbf{Group MAE}}                                            & \multicolumn{3}{c}{\textbf{Overall}}          \\ \cline{2-15} 
		& \textbf{MAE0} & \textbf{MAE1} & \textbf{MAE2} & \multicolumn{1}{c|}{\textbf{MAE3}} & \textbf{MAE$\downarrow$}  & $\sigma\downarrow$             & \textbf{AAR$\uparrow$}  & \textbf{MAE0} & \textbf{MAE1} & \textbf{MAE2} & \multicolumn{1}{c|}{\textbf{MAE3}} & \textbf{MAE$\downarrow$}  & $\sigma\downarrow$             & \textbf{AAR$\uparrow$}  \\ \hline
		Baseline                         & -             & 3.21          & 2.16          & \multicolumn{1}{c|}{5.23}          & 2.42          & 1.69          & 5.89          & -             & 3.14          & 2.29          & \multicolumn{1}{c|}{4.97}          & 2.50          & 1.47          & 6.03          \\
		\cellcolor{gray!15}\textbf{GroupFace (Ours)}               & \cellcolor{gray!15}\textbf{-}    & \cellcolor{gray!15}\textbf{2.17} & \cellcolor{gray!15}\textbf{2.03} & \multicolumn{1}{c|}{\cellcolor{gray!15}\textbf{4.27}} & \cellcolor{gray!15}\textbf{2.09} & \cellcolor{gray!15}\textbf{1.25} & \cellcolor{gray!15}\textbf{6.66} & \cellcolor{gray!15}\textbf{-}    & \cellcolor{gray!15}\textbf{2.59} & \cellcolor{gray!15}\textbf{2.14} & \multicolumn{1}{c|}{\cellcolor{gray!15}\textbf{4.29}} & \cellcolor{gray!15}\textbf{2.27} & \cellcolor{gray!15}\textbf{1.18} & \cellcolor{gray!15}\textbf{6.55} \\ \hline
		\multirow{3}{*}{\textbf{Method}} & \multicolumn{7}{c|}{\textbf{UTK-Face}}                                                                                             & \multicolumn{7}{c}{\textbf{MIVIA}}                                                                                                 \\ \cline{2-15} 
		& \multicolumn{4}{c|}{\textbf{Group MAE}}                                            & \multicolumn{3}{c|}{\textbf{Overall}}         & \multicolumn{4}{c|}{\textbf{Group MAE}}                                            & \multicolumn{3}{c}{\textbf{Overall}}          \\ \cline{2-15} 
		& \textbf{MAE0} & \textbf{MAE1} & \textbf{MAE2} & \multicolumn{1}{c|}{\textbf{MAE3}} & \textbf{MAE$\downarrow$}  & $\sigma\downarrow$             & \textbf{AAR$\uparrow$}  & \textbf{MAE0} & \textbf{MAE1} & \textbf{MAE2} & \multicolumn{1}{c|}{\textbf{MAE3}} & \textbf{MAE$\downarrow$}  & $\sigma\downarrow$             & \textbf{AAR$\uparrow$}  \\ \hline
		Baseline                         & 5.24          & 8.45          & 4.11          & \multicolumn{1}{c|}{6.22}          & 4.78          & 2.01          & 3.21          & 6.86          & 3.23          & 1.62          & \multicolumn{1}{c|}{3.81}          & 1.86          & 2.77          & 5.37          \\
		\cellcolor{gray!15}\textbf{GroupFace (Ours)}               & \cellcolor{gray!15}\textbf{4.49} & \cellcolor{gray!15}\textbf{5.91} & \cellcolor{gray!15}\textbf{4.08} & \multicolumn{1}{c|}{\cellcolor{gray!15}\textbf{4.63}} & \cellcolor{gray!15}\textbf{4.32} & \cellcolor{gray!15}\textbf{0.82} & \cellcolor{gray!15}\textbf{4.86} & \cellcolor{gray!15}\textbf{3.78} & \cellcolor{gray!15}\textbf{2.31} & \cellcolor{gray!15}\textbf{1.58} & \multicolumn{1}{c|}{\cellcolor{gray!15}\textbf{2.56}} & \cellcolor{gray!15}\textbf{1.68} & \cellcolor{gray!15}\textbf{1.17} & \cellcolor{gray!15}\textbf{7.15} \\ \hline
	\end{tabular}
	\label{table6}	                   
\end{table*}

\subsection{Long-tailed Generalization Analysis (\textbf{RQ2})}
To extra evaluate the generalization performance of our architecture GroupFace in different age groups, we consider evaluating the long-tailed age estimation on three imbalanced datasets. Our experiments categorize the samples into four groups: children (0-12), teenager (13-17), adult (18-65), and senior (66+). In most datasets, adult is treated as the head class, while children, teenager, and senior are treated as the tail classes.

Specifically, as shown in TABLE \ref{table1}, MORPH II with an age span of 16-77, lacks the children group and has a teenager group of about 20$\%$, with the tail class concentrated in senior at about 1.7$\%$. For CACD ranging 16-62 years, it missing the children and senior groups, so these datasets are not used in the long-tailed experiment. In addition, the Chalearn LAP 2015 dataset is also not used in that experiment considering that the data distribution imbalance is relatively insignificant and contains only 7591 images. In contrast, both UTK-Face and MIVIA are large-scale datasets with long age spans and significant long-tailed distributions, which are appropriate for evaluating the imbalanced learning performance of the models. And we perform the generalization analysis on the Morph II dataset without the help of external dateset, while the analysis on the UTK-Face and MIVIA datasets with the help of pre-training on the IMDB-WIKI dataset.

Following the long-tailed recognition \cite{bao2021lae}, we apply both MAE and AAR to evaluate the performance of different age groups. TABLE \ref{table6} shows the long-tailed recognition performance of GroupFace with three different datasets, where Baseline utilizes one-hop GCN and sofmax for classification without margin optimization. It is obvious that our GroupFace not only has a large improvement in the overall MAE but also has a significant improvement in the group MAE. Under two settings of MORPH II, the group MAE gap between our different classes narrows and $\sigma$ decreases significantly, with the AAR reaching as high as 6.64. Meanwhile, our long-tailed recognition performance improvement is more obvious in UTK-Face and MIVIA, where the age span and imbalance are more severe. The lowest $\sigma$ of 0.77 and the highest AAR of 7.54 are achieved in MIVIA. This is due to the effective integration of EMAGCN and dynamic group-aware margin optimization in GroupFace, which greatly improves the model's balanced generalization performance for long-tailed recognition.

\begin{table}[!t]
	\caption{\textcolor{black}{The long-tailed age estimation comparisons on MIVIA dataset.}\textcolor{black}{(‘↓’ indicates the smaller is better, while ‘↑’ indicates the larger the better.)}}
	\renewcommand\arraystretch{1.4}
	\tabcolsep=0.028\linewidth
	\color{black}
	\centering
	\begin{tabular}{c|ccc|ccc}
		\hline
		\multirow{2}{*}{Method} & \multicolumn{3}{c|}{GroupMAE}                 & \multicolumn{3}{c}{Overall}                   \\ \cline{2-7} 
		& 0-17          & 18-65         & 66-100        & MAE$\downarrow$           & $\sigma\downarrow$             & AAR$\uparrow$            \\ \hline
		GLAE                    & 3.41          & 1.68          & \textbf{2.31} & 1.73          & \textbf{0.69} & \textbf{7.58} \\  
		\cellcolor{gray!15} GroupFace               & 	\cellcolor{gray!15} \textbf{2.73} & 	\cellcolor{gray!15} \textbf{1.58} & 	\cellcolor{gray!15} 2.56          & 	\cellcolor{gray!15} \textbf{1.68} &	\cellcolor{gray!15} 1.71          & 	\cellcolor{gray!15} 7.15          \\ \hline
	\end{tabular}
	\label{table7}	                   
\end{table}

\textcolor{black}{
Meanwhile, we compare our GroupFace on the MIVIA dataset using the same evaluation metrics with the imbalanced age estimation benchmark GLAE \cite{bao2023general}. Because of the different age grouping strategies, we follow GLAE to combine children and teenager (0-18) in the comparison. As shown in the TABLE \ref{table7}, our method underperforms GLAE in the comprehensive metric AAR, but achieves a lower overall MAE of 1.68. GLAE utilized the smaller ResNet-18 (11M) or the better ResNet-50 (23M) as the backbone network, and designed Adaptive Routing (AR) to select the appropriate classifiers, which achieved both general and imbalanced age estimation with excellent results. Our GroupFace utilizes a more flexible multi-hop graph model (8M) to extract local and global features, and designs dynamic Group-aware margin optimization for imbalanced learning, which also achieves impressive performance.
}

\begin{table*}[!t]
	\caption{The effect of different key components on tree imbalanced age estimation datasets. \textcolor{black}{(‘↓’ indicates the smaller is better, while ‘↑’ indicates the larger the better.)}}
	\renewcommand\arraystretch{1.4}
	\tabcolsep=0.0152\linewidth
	\centering
	\begin{tabular}{cc|ccc|ccc|ccc|ccc}
		\hline
		\multicolumn{2}{c|}{\textbf{Components}} & \multicolumn{3}{c|}{\textbf{MORPH II (Setting I)}} & \multicolumn{3}{c|}{\textbf{MORPH II (Setting II)}} & \multicolumn{3}{c|}{\textbf{UTK-Face}}        & \multicolumn{3}{c}{\textbf{MIVIA}}                                \\ \hline
		\textbf{EMAGCN}     & \textbf{DGMO}     & \textbf{MAE}    & $\sigma\downarrow$               & \textbf{AAR$\uparrow$}   & \textbf{MAE$\downarrow$}    & $\sigma\downarrow$               & \textbf{AAR$\uparrow$}    & \textbf{MAE$\downarrow$}  & $\sigma\downarrow$             & \textbf{AAR$\uparrow$}  & \textbf{MAE$\downarrow$}  & $\sigma\downarrow$             & \textbf{AAR$\uparrow$}                      \\ \hline
		-                   & -                 & 2.42            & 1.69            & 5.89           & 2.50            & 1.47            & 6.03            & 4.78          & 2.01          & 3.21          & 1.86          & 2.77          & 5.37                              \\
		\textbf{\checkmark}          & -                 & 2.18            & 1.57            & 6.25           & 2.29            & 1.32            & 6.39            & 4.61          & 2.14          & 3.25          & 1.72          & 1.95          & 6.33                              \\
		-                   & \textbf{\checkmark}        & 2.24            & 1.38            & 6.38           & 2.37            & 1.26            & 6.37            & 4.69          & 1.58          & 3.73          & 1.81          & 1.21          & 6.98                              \\
		\cellcolor{gray!15}\textbf{\checkmark}          & \cellcolor{gray!15}\textbf{\checkmark}        & \cellcolor{gray!15}\textbf{2.09}   & \cellcolor{gray!15}\textbf{1.25}   & \cellcolor{gray!15}\textbf{6.66}  & \cellcolor{gray!15}\textbf{2.27}   & \cellcolor{gray!15}\textbf{1.18}   & \cellcolor{gray!15}\textbf{6.55}   & \cellcolor{gray!15}\textbf{4.32} & \cellcolor{gray!15}\textbf{0.82} & \cellcolor{gray!15}\textbf{4.86} & \cellcolor{gray!15}\textbf{1.68} & \cellcolor{gray!15}\textbf{1.17}  &\cellcolor{gray!15}\textbf{7.15} \\ \hline
	\end{tabular}
	\label{table8}	                   
\end{table*}

\begin{table*}[!t]
	\caption{The effect of different margin losses on the UTK-Face dataset. \textcolor{black}{(‘↓’ indicates the smaller is better, while ‘↑’ indicates the larger the better.)}}
	\renewcommand\arraystretch{1.4}
	\tabcolsep=0.036\linewidth
	\centering
	\begin{tabular}{c|cccc|ccc}
		\hline
		\multirow{2}{*}{\textbf{Type}} & \multicolumn{4}{c|}{\textbf{Group MAE}}                       & \multicolumn{3}{c}{\textbf{Overall}}          \\ \cline{2-8} 
		& \textbf{MAE0} & \textbf{MAE1} & \textbf{MAE2} & \textbf{MAE3} & \textbf{MAE$\downarrow$}  & $\sigma\downarrow$             & \textbf{AAR$\uparrow$}  \\ \hline
		Sofmax                         & 4.64          & 8.78          & 4.07          & 5.42          & 4.61          & 2.14          & 3.25          \\
		CosFace\cite{wang2018cosface}                & 4.51          & 7.12          & 4.08          & 5.39          & 4.48          & 1.41          & 4.11          \\
		ArcFace \cite{deng2019arcface}               & 4.48          & 7.09          & 4.07          & 5.41          & 4.47          & 1.40          & 4.12          \\
		ElasticFace-Cos \cite{boutros2022elasticface}       & \color{black}\textbf{4.24}          & 6.43          & 4.08          & 5.37          & 4.39          & 1.14          & 4.47          \\
		ElasticFace-Arc \cite{boutros2022elasticface}       & 4.26          & 6.38          & 4.06          & 5.32          & 4.37          & 1.12          & 4.51          \\
		\cellcolor{gray!15}\textbf{GroupFace (Ours)}             & \cellcolor{gray!15}4.49 & \cellcolor{gray!15}\textbf{5.81} & \cellcolor{gray!15}\textbf{4.05} & \cellcolor{gray!15}\textbf{4.93} & \cellcolor{gray!15}\textbf{4.32} & \cellcolor{gray!15}\textbf{0.82} & \cellcolor{gray!15}\textbf{4.86} \\ \hline
	\end{tabular}
	\label{table9}	                   
\end{table*}

\subsection{Ablation Studies (\textbf{RQ3})}
To validate the effectiveness and robustness of our architecture GroupFace, as well as to explore the performance enhancement of the architecture by different components, we perform a series of ablation studies across multiple datasets. And the evaluation of the Morph II dataset is conducted without the help of external dateset, while the evaluation of UTK-Face and MIVIA datasets is with the help of pre-training on the IMDB-WIKI dataset. First, We roughly divide the architecture into two main components: the Enhanced Multi-hop Attention GCN (EMAGCN) and the Dynamic Group-aware Margin Optimization (DGMO), and conduct initial ablation experiments on these two components. Then the different modules and strategies in these two components are further explored in more detail one by one.

\subsubsection{\textbf{Effect of Key Components}}
To verify the effectiveness of the two key components on the entire age estimation architecture, we set the baseline as a one-hop GCN and sofmax for classification without margin optimization, and then compare the gains after adding different components. As shown in TABLE \ref{table8}, the results show that the improvement of overall MAE using EMAGCN is obvious, while DGMO is able to significantly reduce $\sigma$ and improve AAR. Especially in the case of long age-span imbalanced datasets UTK-Face and MIVIA, combining the two improves the AAR by 1.65 and 1.78, respectively. This manifests that EMAGCN can achieve more robust facial feature extraction and capture the discriminative features of the different groups. DGMO reduces the skewness of feature representations and achieves a balanced performance across different groups. Our GroupFace integrating both can attain joint enhancements in both general and long-tailed age estimation tasks.

\begin{figure}[!t]
	\centering
	\includegraphics[width=0.4\textwidth]{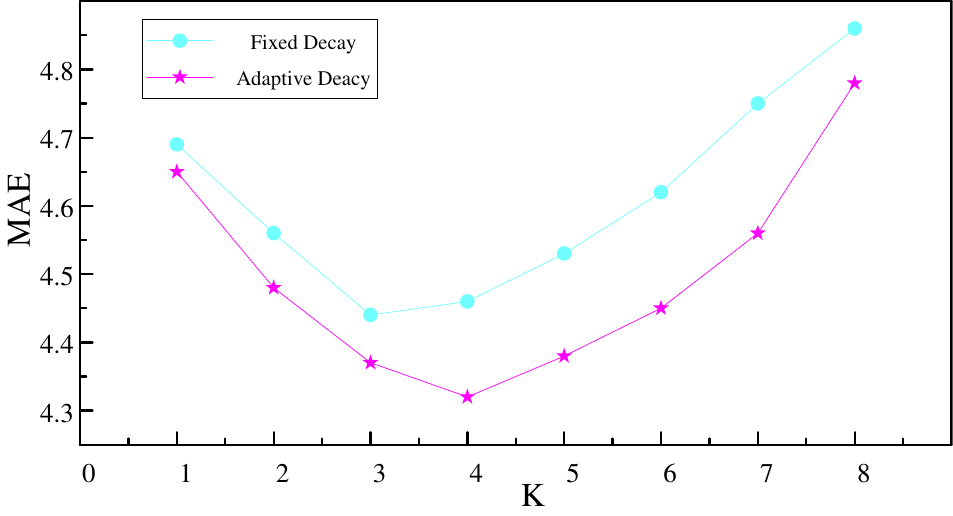}
	\caption{The effect of \textcolor{black}{multi-hop} K and adaptive decay for EMAGCN on UTK-Face dataset.}
	\label{fig_6}
\end{figure}

\begin{figure}[!t]
	\centering
	\includegraphics[width=0.4\textwidth]{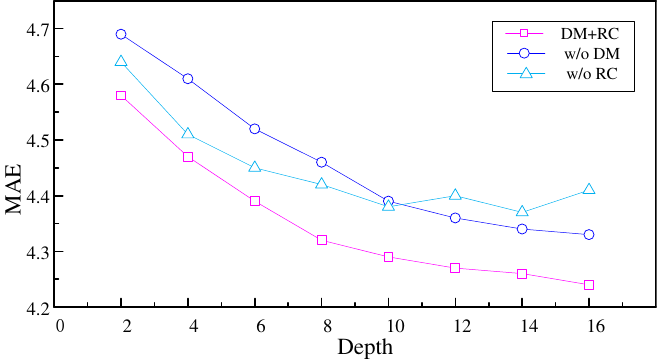}
	\caption{The effect of the designed strategy for EMAGCN and the depth of EMAGCN on the UTK-Face dataset.}
	\label{fig_7}
\end{figure}

\begin{figure}[!t]
	\centering
	\includegraphics[width=0.48\textwidth]{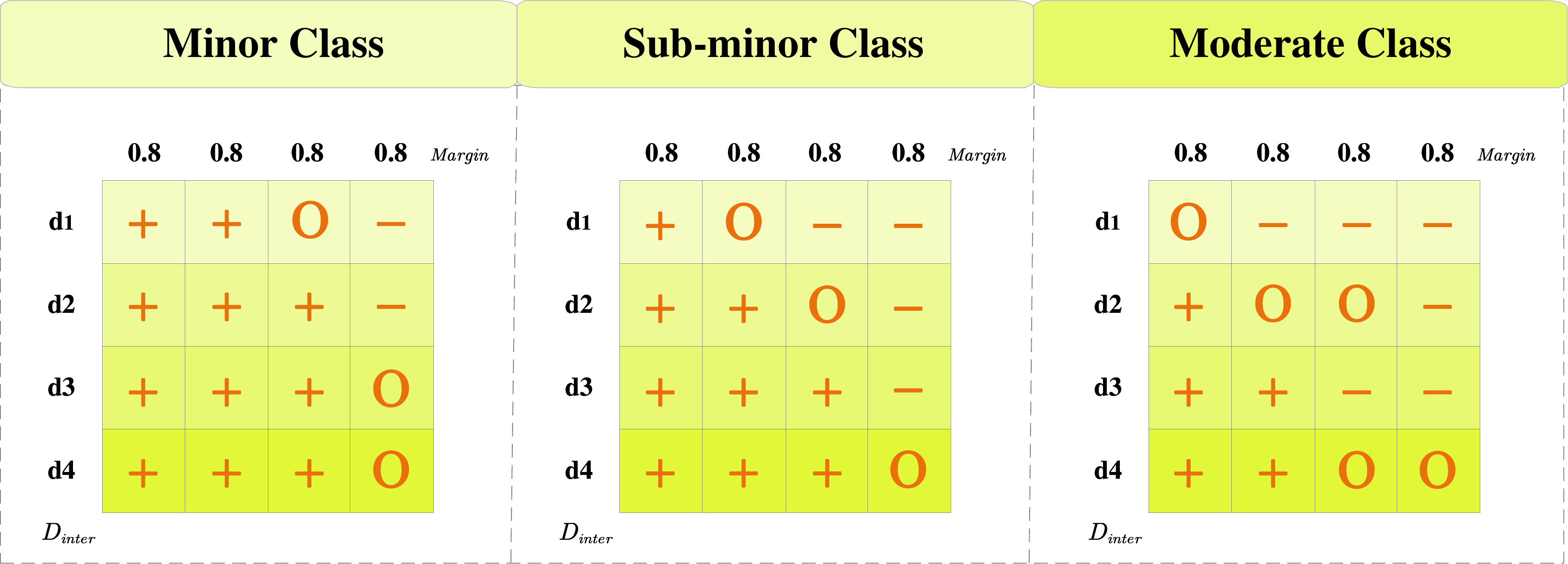}
	\caption{The examples of RL-based group-aware margin strategy from trained agents on the UTK-Face dataset. For the state $s_t=\left\{ G,\,\,D_{inter},\,\,M \right\}$, each grid denotes an action $a_t=\left\{ -1,O,+1 \right\} $ to adjust group margins. $D_{inter}$ is represented as $d_1<d_2<d_3<d_4$, while $M$ is corresponding to discrete spaces $\left\{ 0.2,0.4,0.6,0.8\right\}$.} 
	\label{fig_8}
\end{figure}

\subsubsection{\textbf{Effect of Enhanced Multi-hop Attention GCN}}
To achieve stronger learning of discriminative representations, we design an enhanced multi-hop graph convolutional network to model the aging changes of the face and also design Adaptive Decay strategy (AD), DropMessage (DM), and Residual connection (RC) to enhance the graph model. Before exploring the effects of these designs for deepening the graph model, we adjust the hop count $K$ and the decay on the UTK-Face dataset to find the optimal initial hop count. As shown in Fig. \ref{fig_6}, the MAE is significantly reduced when multi-hop neighbors ($K>1$) are adopted.
\textcolor{black}{
The adaptive decay model reaches the lowest MAE of 4.32 when $K=4$, while the fixed decay model reaches the lowest MAE of 4.44 when $K=3$. Besides, the MAE of adaptive decay is far smaller than fixed decay in most multi-hops, which demonstrates that the adaptive decay strategy can flexibly regulate the degree of attention decay at different distances and improve the effectiveness of multi-hop attention diffusion by reducing noise from long-range distance.} However, the performance gradually goes down as $K$ increases after reaching the \textcolor{black}{minimum}. This may be due to the fact that when the number of hops is too large, aggregating information about neighbors that are too far away introduces more noise than useful information. Therefore, we set the multi-hop count $K$ to 4 and utilize adaptive decay in our experiments.

Based on the optimal hop count $K = 4$, we continue to investigate how much the designed method and strategy improve the multi-hop attention GCN. The comparisons of DM+RC, w/o DM, and w/o RC at different network layers were obtained by removing some components separately. The results are shown in Fig. \ref{fig_7}, DropMessage (DM) helps to keep the diversity of the message delivery while preventing over-fitting, and Residual connection (RC) can maintain initial information while preventing over-smoothing. The combination of the two is more beneficial to keep the performance of the deep network, while with a maximum gap of 0.14 or 0.17 MAE without DM or RC, respectively.

\begin{figure*}[!t]
	\centering
	\includegraphics[width=1\textwidth]{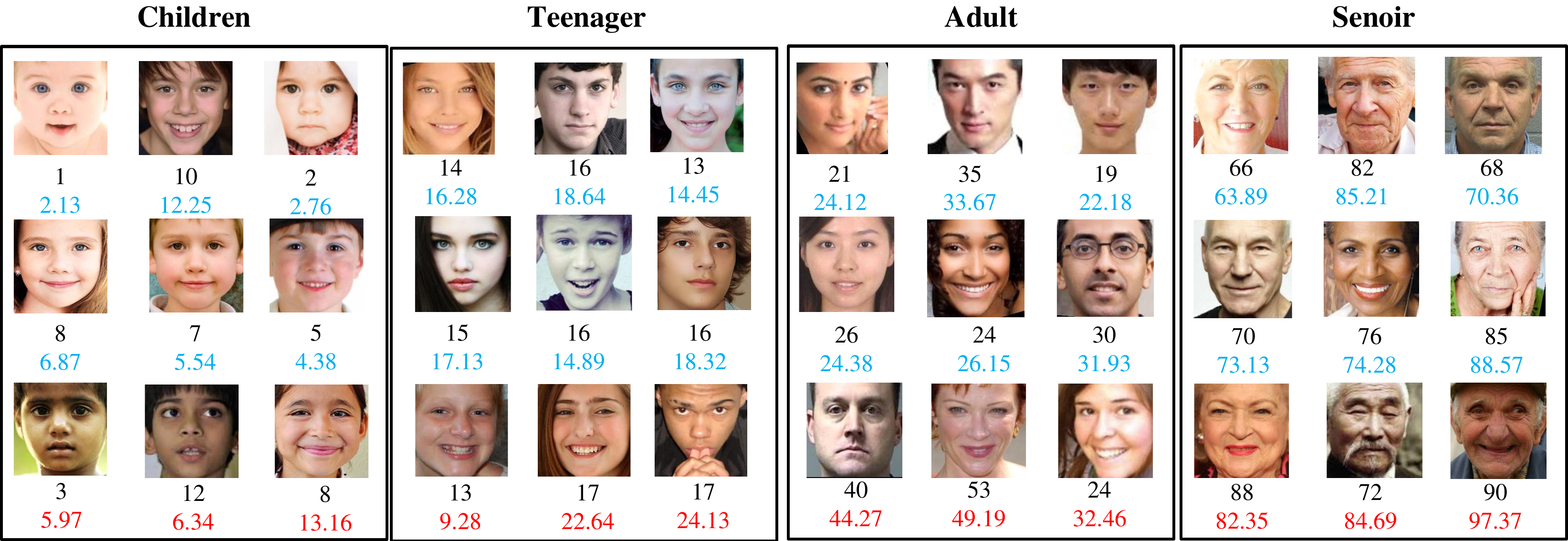}
	\caption{The examples of age estimation results using our GroupFace on UTK-Face dataset. The ground truth label is the black number, the reliable estimation results are shown in the blue number, and the poor estimation results are shown in the red number.}
	\label{fig_9}
\end{figure*}

\subsubsection{\textbf{Effect of Margin Loss Function}}
To investigate the efficacy of our adaptive margin loss on imbalanced age estimation, we compare the experiments using different margin losses on the long-span UTK-Face dataset. As shown in TABLE \ref{table9}, the large margin loss functions CosFace and ArcFace with fixed values can effectively lower the overall MAE compared to the ordinary sofmax. ElasticFace, which can flexibly change the margins, shows significant improvement in the overall performance as well as the performance of different groups. And benefit from the excellent adaptive margin loss function and the dynamic optimization of reinforcement learning, our GroupFace, which can recognize different groups very well, improves the overall recognition accuracy and also effectively achieves a balanced generalization performance. We achieve the lowest overall MAE of 4.32 and the highest AAR of 4.86.

\subsubsection{\textbf{Effect of RL-based Group-aware Margin Optimization}} 
\textcolor{black}{In dynamic group-aware margin strategy, the number of samples of different age groups is used as the classification of head or tail classes, and the head class Adult is regarded as the anchor point. The rest of the classes are regarded as long-tailed classes, which are categorized into Minor Class, Sub-minor Class, and Moderate Class according to the number of samples from smallest to largest. The above categorization varies according to the distribution of age groups in different datasets. In the UTK-Face dataset, the head class is Adult, and the long-tailed classes are Minor Class (Teenager), Sub-minor Class (Senior), and Moderate Class (Children) respectively. The margin of the head class Adult is kept constant after selecting the optimal margin, while the other classes are guided by reinforcement learning to dynamically adjust the margin to minimize the angular skewness between the long-tailed class and the head class. As shown in Fig. \ref{fig_8}, a part of the strategy for different age groups of the trained agent is demonstrated. We can find that the long-tailed classes tend to increase the margin by the head class adult, where the Minor Class (Teenager) increases the largest margin, and the Sub-minor Class (Senior) and Moderate Class (Children) increase the similar margin.}  Having a larger inter-class deviation leads to an increase in the margin, since a larger intra-class distance usually reflects an imbalanced performance in recognizing the group, and thus an increased margin is essential to bolster the group's generalization capability. This demonstrates the adaptability and reliability of our strategy in identifying long-tailed groups, which can dynamically find appropriate margins for different age groups, reducing the skewness of feature representations between different groups and balancing intra-class proximity and inter-class separability.

\subsection{Qualitative Results}
We select the samples of different age groups on the long-span, large-scale UTK-Face dataset for example demonstration of results. As shown in Fig. \ref{fig_9}, our GroupFace performs well in all four age groups and achieves balanced generalization performance in long-tailed recognition. The blue numbers show that our method improves significantly in children, teenager and senior, which is attributed to the mining of discriminative facial features from different age groups, as well as the group-aware margin optimization. The red numbers show some of the failed samples, which may be caused by exaggerated facial expressions, heavy makeup, and so on.

\begin{figure}[!t]
	\centering
	\subfloat[Baseline]{\includegraphics[width=0.23\textwidth]{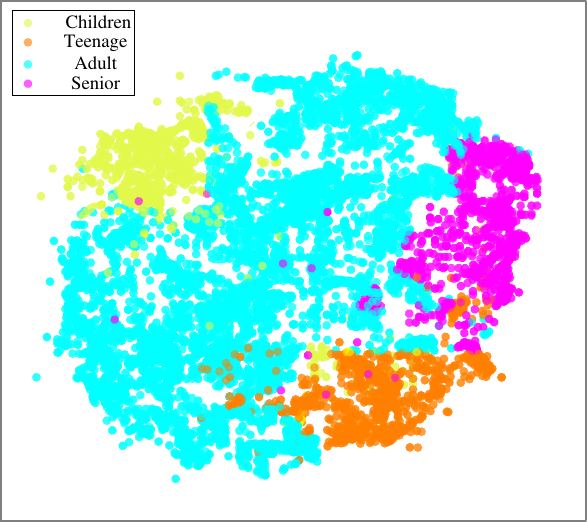}%
		\label{fig10(a)}}
	\hfil
	\subfloat[GroupFace]{\includegraphics[width=0.23\textwidth]{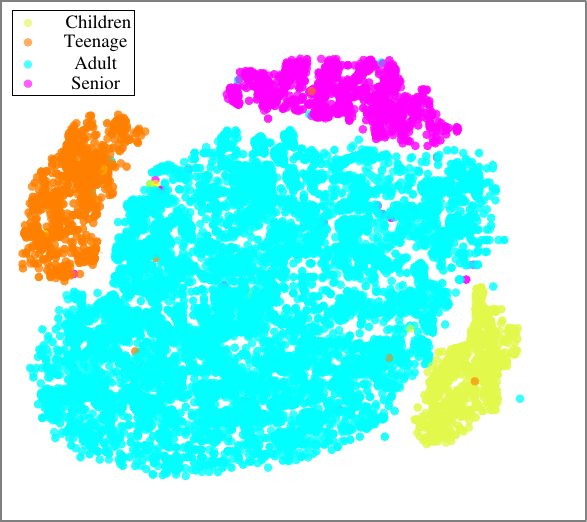}%
		\label{fig10(b)}}
	\caption{The t-SNE visualization of the Baseline and our architecture GroupFace.}
	\label{fig_10}
\end{figure}

\subsection{Feature Visualization}
We further compare the learned feature distributions from Baseline and our GroupFace on UTK-Face dataset by t-SNE visualization. For a fair comparison, Baseline uses a normal one-hop GCN and sofmax for classification without margin optimization. From Fig. \ref{fig10(a)}, it can be seen that the features learned by the Baseline are more dispersed, and the margin overlaps, which is not sufficiently distinguishable for different age groups. While from Fig. \ref{fig10(b)}, it can be observed that our GroupFace has a more compact feature distribution compared to Baseline, and the margins of different age groups are clearly distinguished. This indicates that GroupFace is effective in extracting the discriminative features of different age groups and providing appropriate and unbiased margins for different groups, balancing inter-class separability and intra-class proximity, which also demonstrates that GroupFace is very effective for imbalanced age estimation.

\section{Conclusion}
In this novel, we have presented an innovative collaborative learning framework GroupFace, which integrates an Enhanced Multi-hop Attention Graph Convolutional Network (EMAGCN) and a dynamic group-aware margin strategy based on reinforcement learning. The EMAGCN fuses local and global information to model aging
changes in faces, which can achieve more discriminative representation learning. In addition, the dynamic group-aware margin strategy based on reinforcement learning provides appropriate and unbiased margins for different groups, which can balance inter-class separability and intra-class proximity. Extensive experiments have shown that our GroupFace not only provides a significant improvement in overall estimation accuracy but also balances performance in long-tailed groups. For future work, we are interested in further improving the effectiveness and generalizability of imbalanced learning with the help of the language-image pre-training method.

\section{Acknowledgement}
\textcolor{black}{The author's deepest gratitude goes to the anonymous reviewers and AE for their careful work and thoughtful suggestions that have helped improve this paper substantially.} This work is supported by the National Natural Science Foundation of China (Grant No. 61802444), and the Research Foundation of the Education Bureau of Hunan Province of China (Grant No. 22B0275).

\bibliographystyle{IEEEtran}
\bibliography{reference}

\begin{thebibliography}{10}
\providecommand{\url}[1]{#1}
\csname url@samestyle\endcsname
\providecommand{\newblock}{\relax}
\providecommand{\bibinfo}[2]{#2}
\providecommand{\BIBentrySTDinterwordspacing}{\spaceskip=0pt\relax}
\providecommand{\BIBentryALTinterwordstretchfactor}{4}
\providecommand{\BIBentryALTinterwordspacing}{\spaceskip=\fontdimen2\font plus
\BIBentryALTinterwordstretchfactor\fontdimen3\font minus
  \fontdimen4\font\relax}
\providecommand{\BIBforeignlanguage}[2]{{%
\expandafter\ifx\csname l@#1\endcsname\relax
\typeout{** WARNING: IEEEtran.bst: No hyphenation pattern has been}%
\typeout{** loaded for the language `#1'. Using the pattern for}%
\typeout{** the default language instead.}%
\else
\language=\csname l@#1\endcsname
\fi
#2}}
\providecommand{\BIBdecl}{\relax}
\BIBdecl

\bibitem{rothe2015dex}
R.~Rothe, R.~Timofte, and L.~Van~Gool, ``Dex: Deep expectation of apparent age
  from a single image,'' in \emph{Proceedings of the IEEE international
  conference on computer vision workshops}, 2015, pp. 10--15.

\bibitem{zhang2020ssr}
X.~Zhang, W.~Huang, Q.~Wang, and X.~Li, ``Ssr-net: Spatial--spectral
  reconstruction network for hyperspectral and multispectral image fusion,''
  \emph{IEEE Transactions on Geoscience and Remote Sensing}, vol.~59, no.~7,
  pp. 5953--5965, 2020.

\bibitem{wang2023exploiting}
C.~Wang, Z.~Li, X.~Mo, X.~Tang, and H.~Liu, ``Exploiting unfairness with
  meta-set learning for chronological age estimation,'' \emph{IEEE Transactions
  on Information Forensics and Security}, 2023.

\bibitem{shou2022conversational}
Y.~Shou, T.~Meng, W.~Ai, S.~Yang, and K.~Li, ``Conversational emotion
  recognition studies based on graph convolutional neural networks and a
  dependent syntactic analysis,'' \emph{Neurocomputing}, vol. 501, pp.
  629--639, 2022.

\bibitem{shou2025masked}
Y.~Shou, X.~Cao, H.~Liu, and D.~Meng, ``Masked contrastive graph representation
  learning for age estimation,'' \emph{Pattern Recognition}, vol. 158, p.
  110974, 2025.

\bibitem{shou2022object}
Y.~Shou, T.~Meng, W.~Ai, C.~Xie, H.~Liu, and Y.~Wang, ``Object detection in
  medical images based on hierarchical transformer and mask mechanism,''
  \emph{Computational Intelligence and Neuroscience}, vol. 2022, no.~1, p.
  5863782, 2022.

\bibitem{shou2023comprehensive}
Y.~Shou, T.~Meng, W.~Ai, N.~Yin, and K.~Li, ``A comprehensive survey on
  multi-modal conversational emotion recognition with deep learning,''
  \emph{arXiv preprint arXiv:2312.05735}, 2023.

\bibitem{shou2024adversarial}
Y.~Shou, T.~Meng, W.~Ai, F.~Zhang, N.~Yin, and K.~Li, ``Adversarial alignment
  and graph fusion via information bottleneck for multimodal emotion
  recognition in conversations,'' \emph{Information Fusion}, vol. 112, p.
  102590, 2024.

\bibitem{meng2024deep}
T.~Meng, Y.~Shou, W.~Ai, N.~Yin, and K.~Li, ``Deep imbalanced learning for
  multimodal emotion recognition in conversations,'' \emph{IEEE Transactions on
  Artificial Intelligence}, 2024.

\bibitem{shou2023adversarial}
Y.~Shou, T.~Meng, W.~Ai, N.~Yin, and K.~Li, ``Adversarial representation with
  intra-modal and inter-modal graph contrastive learning for multimodal emotion
  recognition,'' \emph{arXiv preprint arXiv:2312.16778}, 2023.

\bibitem{chen2017using}
S.~Chen, C.~Zhang, M.~Dong, J.~Le, and M.~Rao, ``Using ranking-cnn for age
  estimation,'' in \emph{Proceedings of the IEEE conference on computer vision
  and pattern recognition}, 2017, pp. 5183--5192.

\bibitem{dosovitskiy2020image}
A.~Dosovitskiy, L.~Beyer, A.~Kolesnikov, D.~Weissenborn, X.~Zhai,
  T.~Unterthiner, M.~Dehghani, M.~Minderer, G.~Heigold, S.~Gelly \emph{et~al.},
  ``An image is worth 16x16 words: Transformers for image recognition at
  scale,'' \emph{arXiv preprint arXiv:2010.11929}, 2020.

\bibitem{kuprashevich2023mivolo}
M.~Kuprashevich and I.~Tolstykh, ``Mivolo: Multi-input transformer for age and
  gender estimation,'' in \emph{International Conference on Analysis of Images,
  Social Networks and Texts}.\hskip 1em plus 0.5em minus 0.4em\relax Springer,
  2023, pp. 212--226.

\bibitem{qin2023swinface}
L.~Qin, M.~Wang, C.~Deng, K.~Wang, X.~Chen, J.~Hu, and W.~Deng, ``Swinface: a
  multi-task transformer for face recognition, expression recognition, age
  estimation and attribute estimation,'' \emph{IEEE Transactions on Circuits
  and Systems for Video Technology}, 2023.

\bibitem{ai2023gcn}
W.~Ai, Y.~Shou, T.~Meng, N.~Yin, and K.~Li, ``Der-gcn: Dialogue and event
  relation-aware graph convolutional neural network for multimodal dialogue
  emotion recognition,'' \emph{arXiv preprint arXiv:2312.10579}, 2023.

\bibitem{ai2024gcn}
W.~Ai, Y.~Shou, T.~Meng, and K.~Li, ``Der-gcn: Dialog and event relation-aware
  graph convolutional neural network for multimodal dialog emotion
  recognition,'' \emph{IEEE Transactions on Neural Networks and Learning
  Systems}, 2024.

\bibitem{meng2024multi}
T.~Meng, Y.~Shou, W.~Ai, J.~Du, H.~Liu, and K.~Li, ``A multi-message passing
  framework based on heterogeneous graphs in conversational emotion
  recognition,'' \emph{Neurocomputing}, vol. 569, p. 127109, 2024.

\bibitem{shou2024contrastive}
Y.~Shou, H.~Lan, and X.~Cao, ``Contrastive graph representation learning with
  adversarial cross-view reconstruction and information bottleneck,''
  \emph{arXiv preprint arXiv:2408.00295}, 2024.

\bibitem{shou2024spegcl}
Y.~Shou, X.~Cao, and D.~Meng, ``Spegcl: Self-supervised graph spectrum
  contrastive learning without positive samples,'' \emph{arXiv preprint
  arXiv:2410.10365}, 2024.

\bibitem{shou2024efficient}
Y.~Shou, W.~Ai, J.~Du, T.~Meng, and H.~Liu, ``Efficient long-distance latent
  relation-aware graph neural network for multi-modal emotion recognition in
  conversations,'' \emph{arXiv preprint arXiv:2407.00119}, 2024.

\bibitem{ying2021prediction}
R.~Ying, Y.~Shou, and C.~Liu, ``Prediction model of dow jones index based on
  lstm-adaboost,'' in \emph{2021 International Conference on Communications,
  Information System and Computer Engineering (CISCE)}.\hskip 1em plus 0.5em
  minus 0.4em\relax IEEE, 2021, pp. 808--812.

\bibitem{shou2023graph}
Y.~Shou, W.~Ai, T.~Meng, and N.~Yin, ``Graph information bottleneck for remote
  sensing segmentation,'' \emph{arXiv preprint arXiv:2312.02545}, 2023.

\bibitem{shou2023czl}
Y.~Shou, W.~Ai, T.~Meng, and K.~Li, ``Czl-ciae: Clip-driven zero-shot learning
  for correcting inverse age estimation,'' \emph{arXiv preprint
  arXiv:2312.01758}, 2023.

\bibitem{meng2024masked}
T.~Meng, F.~Zhang, Y.~Shou, H.~Shao, W.~Ai, and K.~Li, ``Masked graph learning
  with recurrent alignment for multimodal emotion recognition in
  conversation,'' \emph{IEEE/ACM Transactions on Audio, Speech, and Language
  Processing}, 2024.

\bibitem{shou2024revisiting}
Y.~Shou, T.~Meng, F.~Zhang, N.~Yin, and K.~Li, ``Revisiting multi-modal emotion
  learning with broad state space models and probability-guidance fusion,''
  \emph{arXiv preprint arXiv:2404.17858}, 2024.

\bibitem{ai2024edge}
W.~Ai, Y.~Wei, H.~Shao, Y.~Shou, T.~Meng, and K.~Li, ``Edge-enhanced
  minimum-margin graph attention network for short text classification,''
  \emph{Expert Systems with Applications}, vol. 251, p. 124069, 2024.

\bibitem{zhang2024multi}
Y.~Zhang, Y.~Shou, T.~Meng, W.~Ai, and K.~Li, ``A multi-view mask contrastive
  learning graph convolutional neural network for age estimation,''
  \emph{Knowledge and Information Systems}, pp. 1--26, 2024.

\bibitem{greco2022effective}
A.~Greco, A.~Saggese, M.~Vento, and V.~Vigilante, ``Effective training of
  convolutional neural networks for age estimation based on knowledge
  distillation,'' \emph{Neural Computing and Applications}, pp. 1--16, 2022.

\bibitem{bao2021lae}
Z.~Bao, Z.~Tan, Y.~Zhu, J.~Wan, X.~Ma, Z.~Lei, and G.~Guo, ``Lae: Long-tailed
  age estimation,'' in \emph{Computer Analysis of Images and Patterns: 19th
  International Conference, CAIP 2021, Virtual Event, September 28--30, 2021,
  Proceedings, Part II 19}.\hskip 1em plus 0.5em minus 0.4em\relax Springer,
  2021, pp. 308--316.

\bibitem{deng2021pml}
Z.~Deng, H.~Liu, Y.~Wang, C.~Wang, Z.~Yu, and X.~Sun, ``Pml: Progressive margin
  loss for long-tailed age classification,'' in \emph{Proceedings of the
  IEEE/CVF conference on computer vision and pattern recognition}, 2021, pp.
  10\,503--10\,512.

\bibitem{cai2022meta}
W.~Cai, X.~Dong, and H.~Liu, ``Meta descent learning for class imbalanced age
  estimation,'' in \emph{2022 IEEE International Conference on Multimedia and
  Expo (ICME)}.\hskip 1em plus 0.5em minus 0.4em\relax IEEE, 2022, pp. 1--6.

\bibitem{ahmad2021graph}
T.~Ahmad, L.~Jin, X.~Zhang, S.~Lai, G.~Tang, and L.~Lin, ``Graph convolutional
  neural network for human action recognition: A comprehensive survey,''
  \emph{IEEE Transactions on Artificial Intelligence}, vol.~2, no.~2, pp.
  128--145, 2021.

\bibitem{yan2024feature}
S.~Yan, C.~Li, H.~Wang, B.~Lin, and Y.~Yuan, ``Feature interactive graph neural
  network for kg-based recommendation,'' \emph{Expert Systems with
  Applications}, vol. 237, p. 121411, 2024.

\bibitem{li2023dynamic}
F.~Li, J.~Feng, H.~Yan, G.~Jin, F.~Yang, F.~Sun, D.~Jin, and Y.~Li, ``Dynamic
  graph convolutional recurrent network for traffic prediction: Benchmark and
  solution,'' \emph{ACM Transactions on Knowledge Discovery from Data},
  vol.~17, no.~1, pp. 1--21, 2023.

\bibitem{kipf2016semi}
T.~N. Kipf and M.~Welling, ``Semi-supervised classification with graph
  convolutional networks,'' \emph{arXiv preprint arXiv:1609.02907}, 2016.

\bibitem{ai2023two}
W.~Ai, F.~Zhang, T.~Meng, Y.~Shou, H.~Shao, and K.~Li, ``A two-stage multimodal
  emotion recognition model based on graph contrastive learning,'' in
  \emph{2023 IEEE 29th International Conference on Parallel and Distributed
  Systems (ICPADS)}.\hskip 1em plus 0.5em minus 0.4em\relax IEEE, 2023, pp.
  397--404.

\bibitem{meng2024revisiting}
T.~Meng, F.~Zhang, Y.~Shou, W.~Ai, N.~Yin, and K.~Li, ``Revisiting multimodal
  emotion recognition in conversation from the perspective of graph spectrum,''
  \emph{arXiv preprint arXiv:2404.17862}, 2024.

\bibitem{ai2024mcsff}
W.~Ai, W.~Deng, H.~Chen, J.~Du, T.~Meng, and Y.~Shou, ``Mcsff: Multi-modal
  consistency and specificity fusion framework for entity alignment,''
  \emph{arXiv preprint arXiv:2410.14584}, 2024.

\bibitem{shou2023graphunet}
Y.~Shou, W.~Ai, T.~Meng, F.~Zhang, and K.~Li, ``Graphunet: Graph make strong
  encoders for remote sensing segmentation,'' in \emph{2023 IEEE 29th
  International Conference on Parallel and Distributed Systems (ICPADS)}.\hskip
  1em plus 0.5em minus 0.4em\relax IEEE, 2023, pp. 2734--2737.

\bibitem{ai2024graph}
W.~Ai, J.~Li, Z.~Wang, J.~Du, T.~Meng, Y.~Shou, and K.~Li, ``Graph contrastive
  learning via cluster-refined negative sampling for semi-supervised text
  classification,'' \emph{arXiv preprint arXiv:2410.18130}, 2024.

\bibitem{shou2024low}
Y.~Shou, H.~Liu, X.~Cao, D.~Meng, and B.~Dong, ``A low-rank matching attention
  based cross-modal feature fusion method for conversational emotion
  recognition,'' \emph{IEEE Transactions on Affective Computing}, 2024.

\bibitem{shou2024graph}
Y.~Shou, P.~Yan, X.~Yuan, X.~Cao, Q.~Zhao, and D.~Meng, ``Graph domain
  adaptation with dual-branch encoder and two-level alignment for whole slide
  image-based survival prediction,'' \emph{arXiv preprint arXiv:2411.14001},
  2024.

\bibitem{ai2024seg}
W.~Ai, Y.~Gao, J.~Li, J.~Du, T.~Meng, Y.~Shou, and K.~Li, ``Seg: Seeds-enhanced
  iterative refinement graph neural network for entity alignment,'' \emph{arXiv
  preprint arXiv:2410.20733}, 2024.

\bibitem{ai2024contrastive}
W.~Ai, J.~Li, Z.~Wang, Y.~Wei, T.~Meng, and K.~Li, ``Contrastive multi-graph
  learning with neighbor hierarchical sifting for semi-supervised text
  classification,'' \emph{Expert Systems with Applications}, p. 125952, 2024.

\bibitem{fu2024sdr}
F.~Fu, W.~Ai, F.~Yang, Y.~Shou, T.~Meng, and K.~Li, ``Sdr-gnn: Spectral domain
  reconstruction graph neural network for incomplete multimodal learning in
  conversational emotion recognition,'' \emph{Knowledge-Based Systems}, p.
  112825, 2024.

\bibitem{shou2024dynamic}
Y.~Shou, T.~Meng, W.~Ai, and K.~Li, ``Dynamic graph neural ordinary
  differential equation network for multi-modal emotion recognition in
  conversation,'' \emph{arXiv preprint arXiv:2412.02935}, 2024.

\bibitem{velivckovic2017graph}
P.~Veli{\v{c}}kovi{\'c}, G.~Cucurull, A.~Casanova, A.~Romero, P.~Lio, and
  Y.~Bengio, ``Graph attention networks,'' \emph{arXiv preprint
  arXiv:1710.10903}, 2017.

\bibitem{hamilton2017inductive}
W.~Hamilton, Z.~Ying, and J.~Leskovec, ``Inductive representation learning on
  large graphs,'' \emph{Advances in neural information processing systems},
  vol.~30, 2017.

\bibitem{wu2019simplifying}
F.~Wu, A.~Souza, T.~Zhang, C.~Fifty, T.~Yu, and K.~Weinberger, ``Simplifying
  graph convolutional networks,'' in \emph{International conference on machine
  learning}.\hskip 1em plus 0.5em minus 0.4em\relax PMLR, 2019, pp. 6861--6871.

\bibitem{abu2019mixhop}
S.~Abu-El-Haija, B.~Perozzi, A.~Kapoor, N.~Alipourfard, K.~Lerman,
  H.~Harutyunyan, G.~Ver~Steeg, and A.~Galstyan, ``Mixhop: Higher-order graph
  convolutional architectures via sparsified neighborhood mixing,'' in
  \emph{international conference on machine learning}.\hskip 1em plus 0.5em
  minus 0.4em\relax PMLR, 2019, pp. 21--29.

\bibitem{wang2020multi}
G.~Wang, R.~Ying, J.~Huang, and J.~Leskovec, ``Multi-hop attention graph neural
  network,'' \emph{arXiv preprint arXiv:2009.14332}, 2020.

\bibitem{zhang2019c3ae}
C.~Zhang, S.~Liu, X.~Xu, and C.~Zhu, ``C3ae: Exploring the limits of compact
  model for age estimation,'' in \emph{Proceedings of the IEEE/CVF conference
  on computer vision and pattern recognition}, 2019, pp. 12\,587--12\,596.

\bibitem{shin2022moving}
N.-H. Shin, S.-H. Lee, and C.-S. Kim, ``Moving window regression: A novel
  approach to ordinal regression,'' in \emph{Proceedings of the IEEE/CVF
  conference on computer vision and pattern recognition}, 2022, pp.
  18\,760--18\,769.

\bibitem{chen2023daa}
P.~Chen, X.~Zhang, Y.~Li, J.~Tao, B.~Xiao, B.~Wang, and Z.~Jiang, ``Daa: A
  delta age adain operation for age estimation via binary code transformer,''
  in \emph{Proceedings of the IEEE/CVF Conference on Computer Vision and
  Pattern Recognition}, 2023, pp. 15\,836--15\,845.

\bibitem{xiong2024deep}
H.~Xiong and A.~Yao, ``Deep imbalanced regression via hierarchical
  classification adjustment,'' in \emph{Proceedings of the IEEE/CVF Conference
  on Computer Vision and Pattern Recognition}, 2024, pp. 23\,721--23\,730.

\bibitem{li2022nested}
J.~Li, Z.~Tan, J.~Wan, Z.~Lei, and G.~Guo, ``Nested collaborative learning for
  long-tailed visual recognition,'' in \emph{Proceedings of the IEEE/CVF
  Conference on Computer Vision and Pattern Recognition}, 2022, pp. 6949--6958.

\bibitem{baik2024dbn}
J.~S. Baik, I.~Y. Yoon, and J.~W. Choi, ``Dbn-mix: Training dual branch network
  using bilateral mixup augmentation for long-tailed visual recognition,''
  \emph{Pattern Recognition}, vol. 147, p. 110107, 2024.

\bibitem{wang2020mitigating}
M.~Wang and W.~Deng, ``Mitigating bias in face recognition using skewness-aware
  reinforcement learning,'' in \emph{Proceedings of the IEEE/CVF conference on
  computer vision and pattern recognition}, 2020, pp. 9322--9331.

\bibitem{bao2023general}
Z.~Bao, Z.~Tan, J.~Li, J.~Wan, X.~Ma, and Z.~Lei, ``General vs. long-tailed age
  estimation: An approach to kill two birds with one stone,'' \emph{IEEE
  Transactions on Image Processing}, vol.~32, pp. 6155--6167, 2023.

\bibitem{lin2020deep}
E.~Lin, Q.~Chen, and X.~Qi, ``Deep reinforcement learning for imbalanced
  classification,'' \emph{Applied Intelligence}, vol.~50, no.~8, pp.
  2488--2502, 2020.

\bibitem{liu2019fair}
B.~Liu, W.~Deng, Y.~Zhong, M.~Wang, J.~Hu, X.~Tao, and Y.~Huang, ``Fair loss:
  Margin-aware reinforcement learning for deep face recognition,'' in
  \emph{Proceedings of the IEEE/CVF international conference on computer
  vision}, 2019, pp. 10\,052--10\,061.

\bibitem{han2022vision}
K.~Han, Y.~Wang, J.~Guo, Y.~Tang, and E.~Wu, ``Vision gnn: An image is worth
  graph of nodes,'' \emph{Advances in neural information processing systems},
  vol.~35, pp. 8291--8303, 2022.

\bibitem{fang2023dropmessage}
T.~Fang, Z.~Xiao, C.~Wang, J.~Xu, X.~Yang, and Y.~Yang, ``Dropmessage: Unifying
  random dropping for graph neural networks,'' in \emph{Proceedings of the AAAI
  Conference on Artificial Intelligence}, vol.~37, no.~4, 2023, pp. 4267--4275.

\bibitem{wang2018cosface}
H.~Wang, Y.~Wang, Z.~Zhou, X.~Ji, D.~Gong, J.~Zhou, Z.~Li, and W.~Liu,
  ``Cosface: Large margin cosine loss for deep face recognition,'' in
  \emph{Proceedings of the IEEE conference on computer vision and pattern
  recognition}, 2018, pp. 5265--5274.

\bibitem{deng2019arcface}
J.~Deng, J.~Guo, N.~Xue, and S.~Zafeiriou, ``Arcface: Additive angular margin
  loss for deep face recognition,'' in \emph{Proceedings of the IEEE/CVF
  conference on computer vision and pattern recognition}, 2019, pp. 4690--4699.

\bibitem{boutros2022elasticface}
F.~Boutros, N.~Damer, F.~Kirchbuchner, and A.~Kuijper, ``Elasticface: Elastic
  margin loss for deep face recognition,'' in \emph{Proceedings of the IEEE/CVF
  conference on computer vision and pattern recognition}, 2022, pp. 1578--1587.

\bibitem{xu2024x2}
J.~Xu, X.~Liu, X.~Zhang, Y.-W. Si, X.~Li, Z.~Shi, K.~Wang, and X.~Gong,
  ``X2-softmax: Margin adaptive loss function for face recognition,''
  \emph{Expert Systems with Applications}, p. 123791, 2024.

\bibitem{ricanek2006morph}
K.~Ricanek and T.~Tesafaye, ``Morph: A longitudinal image database of normal
  adult age-progression,'' in \emph{7th international conference on automatic
  face and gesture recognition (FGR06)}.\hskip 1em plus 0.5em minus 0.4em\relax
  IEEE, 2006, pp. 341--345.

\bibitem{gao2018age}
B.-B. Gao, H.-Y. Zhou, J.~Wu, and X.~Geng, ``Age estimation using expectation
  of label distribution learning.'' in \emph{IJCAI}, vol.~1, 2018, p.~3.

\bibitem{tan2017efficient}
Z.~Tan, J.~Wan, Z.~Lei, R.~Zhi, G.~Guo, and S.~Z. Li, ``Efficient group-n
  encoding and decoding for facial age estimation,'' \emph{IEEE transactions on
  pattern analysis and machine intelligence}, vol.~40, no.~11, pp. 2610--2623,
  2017.

\bibitem{zhang2017age}
Z.~Zhang, Y.~Song, and H.~Qi, ``Age progression/regression by conditional
  adversarial autoencoder,'' in \emph{Proceedings of the IEEE conference on
  computer vision and pattern recognition}, 2017, pp. 5810--5818.

\bibitem{escalera2015chalearn}
S.~Escalera, J.~Fabian, P.~Pardo, X.~Bar{\'o}, J.~Gonzalez, H.~J. Escalante,
  D.~Misevic, U.~Steiner, and I.~Guyon, ``Chalearn looking at people 2015:
  Apparent age and cultural event recognition datasets and results,'' in
  \emph{Proceedings of the IEEE international conference on computer vision
  workshops}, 2015, pp. 1--9.

\bibitem{chen2015face}
B.-C. Chen, C.-S. Chen, and W.~H. Hsu, ``Face recognition and retrieval using
  cross-age reference coding with cross-age celebrity dataset,'' \emph{IEEE
  Transactions on Multimedia}, vol.~17, no.~6, pp. 804--815, 2015.

\bibitem{agbo2021deep}
O.~Agbo-Ajala and S.~Viriri, ``Deep learning approach for facial age
  classification: a survey of the state-of-the-art,'' \emph{Artificial
  Intelligence Review}, vol.~54, no.~1, pp. 179--213, 2021.

\bibitem{zhang2016joint}
K.~Zhang, Z.~Zhang, Z.~Li, and Y.~Qiao, ``Joint face detection and alignment
  using multitask cascaded convolutional networks,'' \emph{IEEE signal
  processing letters}, vol.~23, no.~10, pp. 1499--1503, 2016.

\bibitem{kingma2014adam}
D.~P. Kingma and J.~Ba, ``Adam: A method for stochastic optimization,''
  \emph{arXiv preprint arXiv:1412.6980}, 2014.

\bibitem{bao2022divergence}
Z.~Bao, Z.~Tan, J.~Wan, X.~Ma, G.~Guo, and Z.~Lei, ``Divergence-driven
  consistency training for semi-supervised facial age estimation,'' \emph{IEEE
  Transactions on Information Forensics and Security}, vol.~18, pp. 221--232,
  2022.

\bibitem{korban2023taa}
M.~Korban, P.~Youngs, and S.~T. Acton, ``Taa-gcn: A temporally aware adaptive
  graph convolutional network for age estimation,'' \emph{Pattern Recognition},
  vol. 134, p. 109066, 2023.

\bibitem{li2022metaage}
W.~Li, J.~Lu, A.~Wuerkaixi, J.~Feng, and J.~Zhou, ``Metaage: meta-learning
  personalized age estimators,'' \emph{IEEE Transactions on Image Processing},
  vol.~31, pp. 4761--4775, 2022.

\bibitem{shen2017label}
W.~Shen, K.~Zhao, Y.~Guo, and A.~L. Yuille, ``Label distribution learning
  forests,'' \emph{Advances in neural information processing systems}, vol.~30,
  2017.

\bibitem{pan2018mean}
H.~Pan, H.~Han, S.~Shan, and X.~Chen, ``Mean-variance loss for deep age
  estimation from a face,'' in \emph{Proceedings of the IEEE conference on
  computer vision and pattern recognition}, 2018, pp. 5285--5294.

\bibitem{yang2018ssr}
T.-Y. Yang, Y.-H. Huang, Y.-Y. Lin, P.-C. Hsiu, and Y.-Y. Chuang, ``Ssr-net: A
  compact soft stagewise regression network for age estimation.'' in
  \emph{IJCAI}, vol.~5, no.~6, 2018, p.~7.

\bibitem{li2019bridgenet}
W.~Li, J.~Lu, J.~Feng, C.~Xu, J.~Zhou, and Q.~Tian, ``Bridgenet: A
  continuity-aware probabilistic network for age estimation,'' in
  \emph{Proceedings of the IEEE/CVF conference on computer vision and pattern
  recognition}, 2019, pp. 1145--1154.

\bibitem{savchenko2019efficient}
A.~V. Savchenko, ``Efficient facial representations for age, gender and
  identity recognition in organizing photo albums using multi-output convnet,''
  \emph{PeerJ Computer Science}, vol.~5, p. e197, 2019.

\bibitem{cao2020rank}
W.~Cao, V.~Mirjalili, and S.~Raschka, ``Rank consistent ordinal regression for
  neural networks with application to age estimation,'' \emph{Pattern
  Recognition Letters}, vol. 140, pp. 325--331, 2020.

\bibitem{sun2021deep}
H.~Sun, H.~Pan, H.~Han, and S.~Shan, ``Deep conditional distribution learning
  for age estimation,'' \emph{IEEE Transactions on Information Forensics and
  Security}, vol.~16, pp. 4679--4690, 2021.

\bibitem{tan2019deeply}
Z.~Tan, Y.~Yang, J.~Wan, G.~Guo, and S.~Z. Li, ``Deeply-learned hybrid
  representations for facial age estimation.'' in \emph{IJCAI}, 2019, pp.
  3548--3554.

\bibitem{gao2017deep}
B.-B. Gao, C.~Xing, C.-W. Xie, J.~Wu, and X.~Geng, ``Deep label distribution
  learning with label ambiguity,'' \emph{IEEE Transactions on Image
  Processing}, vol.~26, no.~6, pp. 2825--2838, 2017.

\bibitem{liu2020facial}
N.~Liu, F.~Zhang, and F.~Duan, ``Facial age estimation using a multi-task
  network combining classification and regression,'' \emph{IEEE Access},
  vol.~8, pp. 92\,441--92\,451, 2020.

\end{thebibliography}

\vfill

\end{document}